\documentclass[runningheads]{llncs}
\usepackage{graphicx}
\usepackage{tikz}
\usepackage{comment}
\usepackage{amsmath,amssymb} % define this before the line numbering.
\usepackage{color}
\usepackage[accsupp]{axessibility}  % Improves PDF readability for those with disabilities.
\usepackage{booktabs}
\usepackage{esvect}
\usepackage{algorithm}
\usepackage{algpseudocode}
\usepackage{makecell}
\usepackage{pifont}
\usepackage{bm}
\usepackage{multirow}
\usepackage{xspace}
\usepackage{enumitem}
\usepackage{romannum}
\usepackage{wrapfig}

\usepackage{subfigure}
\usepackage{tabularx}

\usepackage{hyperref}
\hypersetup{colorlinks,allcolors=black}

\newcommand{\cmark}{\ding{51}}
\newcommand{\xmark}{\ding{55}}
\newcommand{\cxmark}{\textcolor{black}{\ding{51}}{\small\textcolor{black}{\kern-0.7em\ding{55}}}}

\newcommand{\method}{SecretGen\xspace}
\usepackage{xcolor,colortbl}

\newcommand{\eg}[0]{\textit{e.g.}}
\newcommand{\ie}[0]{\textit{i.e.}}

\usepackage{soul}
\usepackage[normalem]{ulem}
\renewcommand\sout{\bgroup\markoverwith
{\color{orange!90!black}{\rule[.5ex]{2pt}{1pt}}}\ULon}

\definecolor{algocolor}{RGB}{68,114,196}

\usepackage[capitalize]{cleveref}
\crefname{section}{Sec.}{Secs.}
\Crefname{section}{Section}{Sections}
\Crefname{table}{Table}{Tables}
\crefname{table}{Tab.}{Tabs.}

\begin{document}
% \renewcommand\thelinenumber{\color[rgb]{0.2,0.5,0.8}\normalfont\sffamily\scriptsize\arabic{linenumber}\color[rgb]{0,0,0}}
% \renewcommand\makeLineNumber {\hss\thelinenumber\ \hspace{6mm} \rlap{\hskip\textwidth\ \hspace{6.5mm}\thelinenumber}}
% \linenumbers
\pagestyle{headings}
\mainmatter
\def\ECCVSubNumber{6086}  % Insert your submission number here

\title{\method: Privacy Recovery on Pre-Trained Models via Distribution Discrimination} % Replace with your title

% INITIAL SUBMISSION 
\begin{comment}
\titlerunning{ECCV-22 submission ID \ECCVSubNumber} 
\authorrunning{ECCV-22 submission ID \ECCVSubNumber} 
\author{Anonymous ECCV submission}
\institute{Paper ID \ECCVSubNumber}
\end{comment}
%***********************

% CAMERA READY SUBMISSION
%\begin{comment}
\titlerunning{SecretGen: Privacy Recovery on Pre-Trained Models}
% If the paper title is too long for the running head, you can set
% an abbreviated paper title here
%
\author{Zhuowen Yuan\inst{1} \and
Fan Wu\inst{1} \and
Yunhui Long\inst{1} \and
Chaowei Xiao\inst{2} \and
Bo Li\inst{1}}
\authorrunning{Z. Yuan et al.}
% First names are abbreviated in the running head.
% If there are more than two authors, 'et al.' is used.
%
\institute{University of Illinois Urbana-Champaign \and Arizona State University}

% \email{\{zhuowen3,fanw6,lbo\}@illinois.edu, yunhuilong2014@gmail.com, }
%\end{comment}
%******************

\maketitle

\begin{abstract}
    Transfer learning through the use of pre-trained models has become a growing trend for the machine learning community. Consequently, numerous pre-trained models are released online to facilitate further research. However, it raises extensive concerns on whether these pre-trained models would leak privacy-sensitive information of their training data. Thus, in this work, we aim to answer the following questions: ``Can we effectively recover private information from these pre-trained models? What are the sufficient conditions to retrieve such sensitive information?” 
    We first explore different statistical information which can discriminate the private training distribution from other distributions. Based on our observations, we propose a novel private data reconstruction framework, \method, to effectively recover private information. Compared with previous methods which can recover private data with the ground truth label of the targeted recovery instance, \method does not require such prior knowledge, making it more practical. 
    We conduct extensive experiments on different datasets under diverse scenarios to compare \method with other baselines and provide a systematic benchmark to better understand the impact of different auxiliary information and optimization operations. 
    We show that without prior knowledge about true class prediction, \method is able to recover private data with similar performance compared with the ones that leverage such prior knowledge. If the prior knowledge is given, \method will significantly outperform baseline methods. We also propose several quantitative metrics to further quantify the privacy vulnerability of pre-trained models, which will help the model selection for privacy-sensitive applications. Our code is available at: \url{https://github.com/AI-secure/SecretGen}.

\keywords{Privacy; Pre-trained models; Transfer learning}
\end{abstract}

\section{Introduction}
As machine learning has achieved great successes in different domains, such as robotics~\cite{wang2018look}, audio recognition~\cite{conneau2020unsupervised}, and face recognition~\cite{he2016deep}, how to train the learning models efficiently given the available large-scale dataset becomes a timely problem.
Transfer learning, which focuses on transferring knowledge across domains, is a promising learning paradigm~\cite{bengio2012deep}. In particular, many pre-trained models are available currently, such as TensorFlow Hubs~\cite{abadi2016tensorflow} and PyTorch Hubs~\cite{paszke2017automatic}, which can be flexibly used for fine-tuning later for different downstream tasks. As a result, the training paradigm with transfer learning has enabled efficient usage of the large-scale dataset without requiring training every model from scratch. 

However, such an efficient transfer learning paradigm also leads to additional \textit{privacy concerns}. 
For instance, if the training data of the pre-trained models contain privacy-sensitive information, an adversary who downloads the pre-trained models could potentially perform different privacy attacks to infer the private information. 
In particular, membership inference attacks~\cite{leino2020stolen,liu2021encodermi} have been studied to infer whether a private instance is in the training set, and model inversion attacks have been studied to reconstruct the private training instances under certain assumptions~\cite{zhang2020secret,fredrikson2014privacy,fredrikson2015model,yang2019adversarial}, which raises more privacy and safety concerns.

To better understand the privacy vulnerabilities of such pre-trained models, a comprehensive analysis of different types of privacy attacks, especially the severe model inversion attacks, is required. Currently, there are several limitations of existing privacy model inversion attacks. 
First, the current \textit{state-of-the-art} model inversion attack (i.e., GMI)~\cite{zhang2020secret} requires the ground truth label of the reconstructed instances, which is less practical. Furthermore, it is a known challenging problem to label the generated instances based on GANs~\cite{goodfellow2020generative}. Second, many existing model inversion attacks require whitebox access to the target pre-trained model, making it less practical in real-world applications.
% which is not practical hard to achieve in real-world applications.
Thus, in this paper, we mainly aim to ask: \textit{Can we reconstruct private sensitive training instances without requiring such information?
% What are the principles to evaluate the recovered private instances?
}

\looseness=-1
To answer it,  we propose a general private data recovery framework \method, which consists of a generation backbone, a pseudo label predictor, and a latent vector selector. 
We first use a pseudo label predictor to generate a pseudo label for each private instance. Specifically, we randomly sample latent vectors and feed them into the generation backbone to get recovered instances. To stabilize prediction quality, we apply different transformations (\eg cutouts) to such instances before feeding them into the targeted model to get the final predicted pseudo labels.
% The pseudo label predictor  selects the consistent prediction under different transformations and sampled latent vectors as the label prediction, to replace the ground truth prediction assumption. 
% We first explore different statistical information to discriminate public and private data, based on which we propose a pseudo label predictor, which aims to select the consistent prediction under different transformations and sampled latent vectors as the label prediction, to replace the ground truth prediction assumption. 
We then propose a latent vector selector via a proposed selection algorithm to further optimize and constrain the recovery space. 
Finally, we perform joint optimization to train the end-to-end framework as shown in Fig.~\ref{fig:pipeline}.

We conduct comprehensive experiments to evaluate the proposed \method compared with multiple baselines. We show that \method significantly outperforms baselines given the same ground truth label. Even without such information, \method still achieves comparable performance compared to baselines which leverage the ground truth label information. 
In addition, to evaluate the performance of recovered data on downstream tasks, we propose different evaluation protocols considering different usage of the recovered private data, and we show that our observations are consistent for different protocols.
We also evaluate the robustness of \method against the purification defense \cite{yang2020defending}.
Finally, we perform different ablation studies to show the effectiveness of our design choices.
We make the following {\bf contributions}:

\begin{itemize}
    \item We propose a general private data recovery attack (i.e., model inversion) given a pre-trained model, \method, without requiring the ground truth label as prior knowledge under both whitebox and blackbox settings.
    \item We propose a novel label predictor for the reconstructed instances considering different data transformations and latent vector selection, which can be flexibly used in other frameworks.
    \item We propose different evaluation protocols and metrics for evaluating the pre-trained models against general  model inversion attacks. 
    \item We conduct extensive experiments on different models, including the vision transformer and multiple datasets, to provide a benchmark on model inversion attacks. We show that \method significantly outperforms baselines under different settings.
\end{itemize}

\section{Related Work}
Revealing privacy-sensitive information from a trained model has aroused extensive research interest.
\textit{Membership inference attacks} and \textit{model inversion attacks} are two major categories of such attacks.
In \textit{membership inference attacks} \cite{leino2020stolen,liu2021encodermi}, the adversary aims to decide whether a sample is a member of the training set, while in \textit{model inversion attacks} \cite{zhang2020secret,fredrikson2014privacy,fredrikson2015model,yang2019adversarial}, the adversary attempts to reconstruct the training set under certain assumptions.

\cite{fredrikson2014privacy} was the first to propose model inversion attacks aiming at recovering private training data.
The authors demonstrated that personal genetic markers could be effectively recovered given the output of the model and auxiliary knowledge.
\cite{fredrikson2015model} extended model inversion to more complex models, including shallow neural networks for face recognition.
The recovered data with their proposed method are identified as the original person at a much higher rate than random guessing.
However, the reconstructed images are blurry and not visually recognizable to humans.
\cite{yang2019adversarial} proposed a training-based attack by training an auto-encoder on public data. 
The attack can be performed with \textit{blackbox} accesses to the target model and partial (truncated) model predictions.

More recently, \cite{zhang2020secret} proposed generative model inversion attack (GMI).
The authors distill public knowledge by training a conditional GAN on public data and then solve an optimization problem to maximize the probability of the recovered image for the ground truth class label.
GMI significantly outperforms previous methods in re-identification rate of the recovered data, as well as guaranteeing the recovered data are visually plausible.
However, they still require the ground truth label for the target image and \textit{whitebox} access to the victim model, which is often not accessible to the adversary.
Another recent work distributional model inversion attack (DMI) \cite{chen2021knowledge} recovers the private data distribution for each target class by constructing representative samples. However, DMI does not support recovering every private instance given its non-sensitive version (\textit{i.e.} instance-level model inversion), which is the adversary's goal in our setting.

\section{Methodology}
\label{sec:method}
% In this section we will introduce the detailed structure of \method.
% , and we will evaluate it under different settings in Section~\ref{sec:exp}.
\begin{figure}[ht]
    \centering
    \includegraphics[width=.8\textwidth]{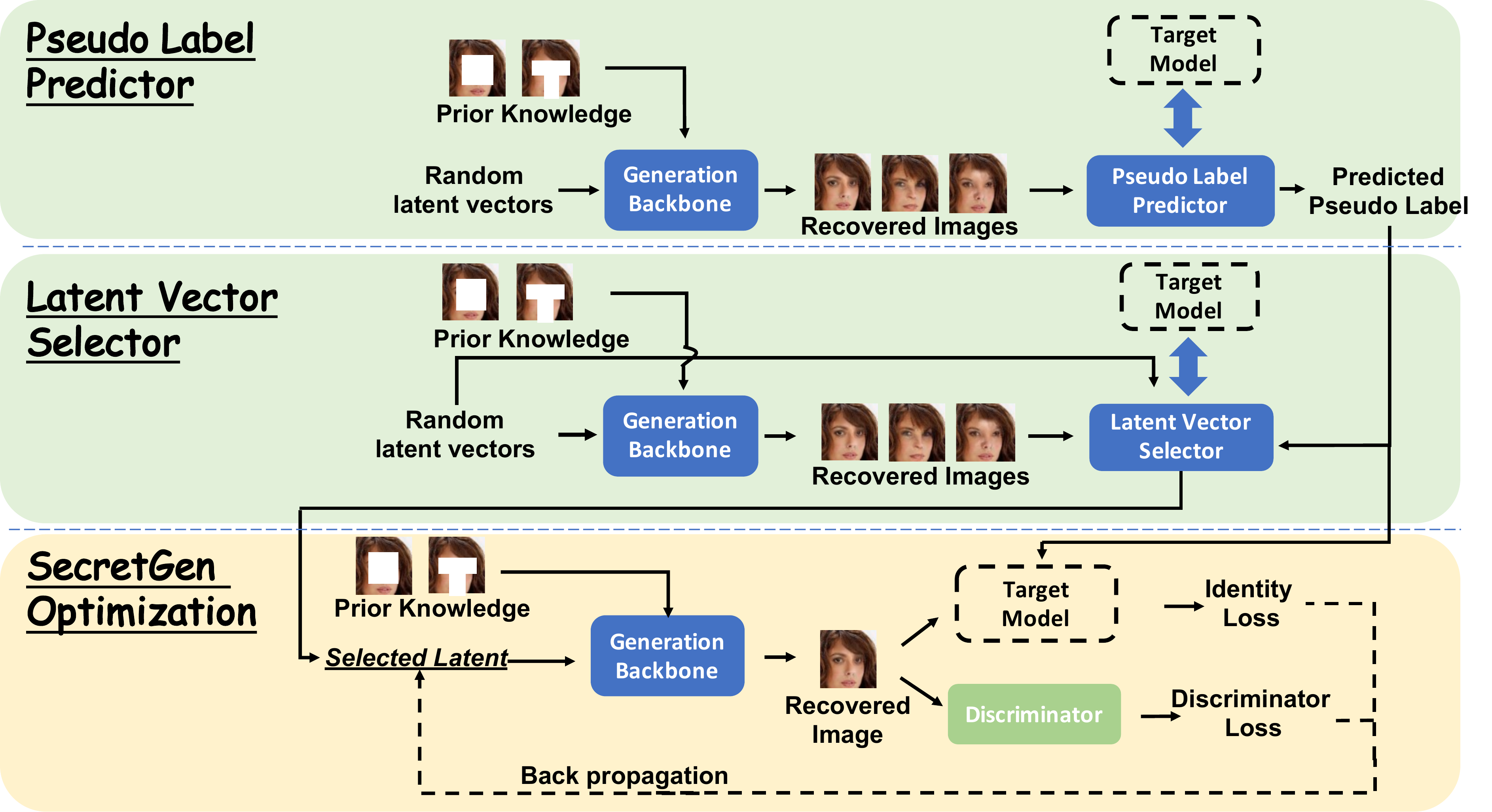}
    \caption{\small Overview of the proposed \method. The \textcolor{algocolor}{\textbf{blue modules}} represent the proposed algorithms. The \textit{Target Model} could allow either whitebox or blackbox access.} 
     
    \label{fig:pipeline}
\end{figure}

\subsection{Problem Formulation}

We focus on recovering the privacy-sensitive training data based on the trained classification models.
Throughout the paper, we will refer to the model that is subjective to attacks as the \textit{target model} {$F$}, which is trained on private training data $D_{\rm pri}^{\rm train}$, aiming to perform evaluation on private test data $D_{\rm pri}^{\rm test}$.
The target model returns a prediction vector $F(x)$ given an input instance $x$.
The prediction vector represents a probability distribution over $C$ classes, where $C$ denotes the number of classes of the whole private dataset $D_{\rm pri} = D_{\rm pri}^{\rm train} \cup D_{\rm pri}^{\rm test}$.
% \fan{Is $D_{\rm pri}=D_{\rm pri}^{\rm train} \cup D_{\rm pri}^{\rm test}$? Maybe first introduce the dataset $D_{\rm pri}$, and then the training and testing set.}
% \fan{change the subscription and superscription to roman font}

The adversary's \textbf{goal} is to recover the private training data $D_{\rm pri}^{\rm train}$ given the trained target model $F$ and certain prior information, \eg, partially corrupted images from $D_{\rm pri}^{\rm train}$. 
% \fan{the gt label may not be a good example of prior information since throughout the paper we assume we don't have access to this info. can directly remove it.}
% \fan{follow the previous order to introduce the details below: first the pre-trained model $F$, and then the other potential info or knowledge such as the corrupted private dataset, the corruption type, and the public dataset. The current structure is unclear.}
% \fan{minor: what's the diff between the term ``prior information'' and the term ``auxiliary knowledge'' (used later)?}
In particular, such corrupted images only contain non-sensitive background information (pixels) $x_{ns}$ with the sensitive region $x_{s}$ cropped out.
These corrupted images are usually easy to obtain, given that such corruption is often applied to protect the privacy of individuals in practice \cite{zhang2020secret}. Specifically, in our evaluation, we consider cropping the whole face using two face datasets, leaving only the non-sensitive background regions (Section~\ref{sec:exp}).
% \yhc{Can we provide some examples for the background pixels we used during the attack to show that they are indeed non-sensitive?}

Regarding the adversary's \textbf{ability}, we consider (1)  \textit{whitebox} access to the target model, where all parameters and intermediate computations of the target model are visible to the adversary, and (2) \textit{blackbox} access to the target model, where the adversary can only obtain the final prediction from the target model $F$.
Additionally, we assume that the adversary also has access to some public data $D_{\rm pub}$ from the similar distribution for general training purpose.

\subsection{Method Overview}

An overview of \method is illustrated in Fig.~\ref{fig:pipeline}, where
\method takes non-sensitive information $x_{ns}$ as input and returns the recovered images that contain privacy-sensitive training information (\eg, human faces).
\method is composed of \textit{three} components: \textit{ generation backbone}, \textit{pseudo label predictor}, and \textit{latent vector selector}, which are jointly optimized under a unified framework.
The \textbf{generation backbone} leverages a conditional GAN trained on public data as a backbone to generate realistic images based on the prior information (\eg, cropped images), and the generation process is controlled by the latent vector $z$ sent to the GAN's generator $G$.
% \fan{here the description for the generation backbone is only how to leverage the trained backbone, while 3.3 is only on training the backbone. disconnected}
The \textbf{pseudo label predictor} predicts the most possible pseudo label for each recovered private image based on the distributional statistics of recovered images.
The \textbf{latent vector selector} selects the optimal latent vector $\hat z$ which is the most likely to contain privacy-sensitive information based on the proposed selection algorithm.
Finally, we perform joint optimization  on the selected $\hat z$, taking the pseudo label provided by the pseudo label predictor as the prediction target, to reconstruct image  $G(\hat z^*, x_{ns})$. In the next following sections, we will describe each component in detail.  
% \fan{should be ``perform optimization on the random variable xxx starting from the initialization $\hat z$}
% with identity loss by leveraging the predicted labels as targets.
% Finally \method returns the generated image $G(\hat z^*, x_{ns})$ from the optimized latent vector $\hat z^*$.
% \fan{in later text (second paragraph in 3.3), $x_{ns}$ explicitly denotes the corrupted public image. use different notations.}
% \fan{maybe change the components to stages (label prediction, latent vector selection, latent vector optimization), since the generation backbone is involved in all stages.}

\subsection{Generation Backbone of \method}
\label{sec:gen-backbone}
To recover the privacy-sensitive training data, we  train a generation backbone for conditional image recovery on public data $D_{\rm pub}$.
In particular, we will start from certain prior knowledge, such as the corrupted private data containing only the nonsensitive information $x_{ns}$.
We then perform the same corruption operation $\mathit{corr}$  on $D_{\rm pub}$ to construct the training set for the generation backbone: $D_{\rm pub\_corr} = \{\mathit{corr}(x)|x\in D_{\rm pub}\} $.

Next, we train a conditional GAN which is composed of two networks: generator $G$ and discriminator $D$.
$G$ is conditioned on $x_{ns} \in D_{\rm pub\_corr}$ and $z$ is the latent vector which is sampled from a prior distribution during training.
Throughout the paper, we use the prior distribution as standard Gaussian distribution.
% ~\yhc{We use the standard Gaussian distribution as the prior distribution?}
We leverage the Wasserstein-GAN loss \cite{gulrajani2017improved} for GAN training:
\begin{equation}
    \min_G\max_{D} \mathcal{L}_\text{wgan} = \mathbb{E}_x[D(x)] - \mathbb{E}_z[D(G(z, x_{ns}))]
\end{equation}

We also incorporate a diversity loss term $\mathcal{L}_{\rm div}$ \cite{yang2019diversity} for training the generator to prevent mode collapsing by sampling different latent vectors, say, $z_1$ and $z_2$:
\begin{equation}
    \mathcal{L}_{\rm div} = -\mathbb{E}_{z_1,z_2}\bigg[\frac{\left\|f(G(z_1, x_{ns}))-f(G(z_2, x_{ns}))\right\|}{\left\|z_1 - z_2\right\|}\bigg]
\label{eq:div}
\end{equation}
where $f$ is the feature extractor of the target model, which returns the feature embeddings of the input images in the \textit{whitebox} setting.
In the \textit{blackbox} setting, we use a feature extractor trained on public data $f_{\rm pub}$ for this process.
The overall loss term for the generator is as following:
\begin{equation}
    \label{eq:gen-backbone}
    \mathcal{L}_{G} = \mathcal{L}_{\rm wgan} + \lambda_{\rm div}\mathcal{L}_{\rm div}
\end{equation}

After the generation backbone is trained, we freeze the parameters for both $G$ and $D$ before we enter the next stage.
We denote $\hat x$ as the recovered image, \ie, $\hat x= G(z, x_{ns})$.
% \fan{The last sentence describes how $G$ can be used later, but I feel it is too brief. Maybe elaborate a bit more on how both $G$ and $D$ can be invoked for what purpose.}

\subsection{Pseudo Label Predictor}
\label{sec:pseudo_label_predictor}
The main challenge in this data reconstruction process is that we have no knowledge about the ground truth label of the private images (related work assumes that they have access to the ground truth label~\cite{zhang2020secret,yang2019adversarial,fredrikson2014privacy,fredrikson2015model}, while we do not).
% \fan{minor: we use both terms ``ground truth prediction'' and ``ground truth label'' in the paper. shall we stick to one? the latter sounds better to me.}
To tackle this problem, we propose a \textit{pseudo label predictor} which infers the label prediction with proposed discrimination metrics. 
We will first introduce the design of our discrimination metric, and then we elaborate on how the pseudo label predictor is optimized.

\noindent\textbf{Discrimination Metric.}\quad
\label{sec:disc-metric}
Given the certain prior knowledge $x_{ns}$, we randomly sample $n$ latent vectors $\{z_i\}_{i=1}^{n}$ from the prior distribution.
We generate $n$ recovered images using our generation backbone: $\{\hat x_i\}_{i=1}^{n}$, where $\hat x_i=G(z_i, x_{ns})$.
% \sout{We further perform $m$ transformations $t_1, ..., t_m$ separately on each of the recovered images.}
In order to improve the prediction stability, we consider prediction under different transformations.
Concretely, let the list of considered transformation functions be $\mathcal {T} = \{t_i\}_{i=1}^{m}$.
On each recovered image $\hat x_i$, we  perform $m$ transformations independently to obtain $m$ transformed images $\{\tilde x_i^j\}_{j=1}^{m}$, where $\tilde x_i^j=t_j(\hat x_i)$.
We additionally define $\tilde x_i^0=\hat x_i$.
Let $F_c(\cdot)$ denote the model's prediction confidence for class label $c$ based on target model $F$.
% We define $\mathcal{M}_1$ as the average confidence for $c$ over the $n$ prediction vectors:
% \begin{equation}
%     \mathcal{M}_1(l) \triangleq \frac{1}{n}\sum_{i=1}^nF_l(\hat x_i)
%     \label{eq:m1}
% \end{equation}
We define the discrimination metric $\mathcal{M}$ on label $c$ as follows:
% \begin{equation}
% \label{eq:m2}
%     \mathcal{M}(c) \triangleq \frac{1}{n(m+1)}\sum_{i=1}^n(F_c(\hat x_i)+\sum_{j=1}^m t_j(F_c(\hat x_i)))
% \end{equation}
% \fan{
% or the following form using the new notations I introduced above
\begin{equation}
\label{eq:metric}
    \mathcal{M}(c;n,m) \triangleq \frac{1}{n(m+1)}\sum_{i=1}^n\sum_{j=0}^{m}F_c(\tilde x_i^j),\quad \forall c \in [1,C].
\end{equation}
% }

The discrimination metric returns a score indicating how likely it is for a label $c$ to be the consistent prediction across different transformations. Based on existing studies of contrastive learning~\cite{chen2020simple}, we will select the class $c$ with the highest discrimination metric score as the final label prediction.

\begin{wrapfigure}{r}{0.5\textwidth}
    \centering
     
    \includegraphics[width=0.48\textwidth]{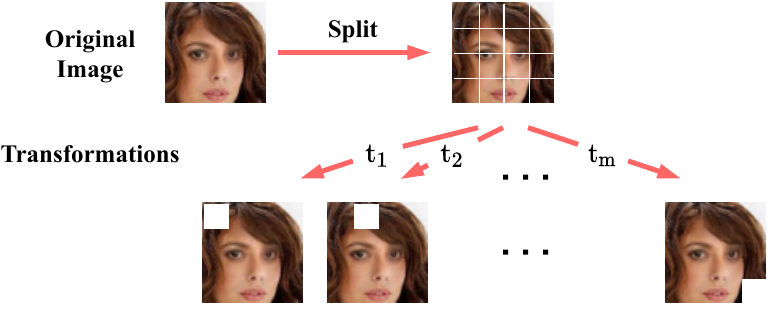}
    \caption{\small Sequential cutout for the recovered image as transformations. The image is first split into $m$ fix-sized patches. Operations of cutting out each patch are viewed as transformations respectively.
    % \fan{For space consideration, maybe we can use wrapfigure to resize this figure as well as Table 1 to half line width. The equations later can also be made smaller.}
    }
     
    \label{fig:cutout}
\end{wrapfigure}

In particular, we define the list of transformations as a sequence of fix-sized cutouts.
We split an image into fix-sized patches and define $t_j$ as the transformation that cuts out the $j$-th patch of the given image, as illustrated in \cref{fig:cutout}.

% Intuitively, we consider the consistent predictions under different transformed recoveries $\{\tilde x_i^j\}_{j=1}^{m}$ as the final prediction, since the correct recoveries should be resistant to the transformations. In addition, the non-training data are usually close to the decision boundary and thus less resilient~\cite{choquette2021label}.~\yhc{Does the following paragraph explains the message we want to convey here?}

Intuitively, the discrimination metric $\mathcal{M}(c;n,m)$ should preserve the following properties.
First, $\mathcal{M}(c;n,m)$ is likely to have a higher score when $c$ equals the label associated with the corrupted image $x_{ns}$ since the model has learned some correlation between the non-sensitive background information in $x_{ns}$ and the label of the original image. 
Such correlation should be stronger if the target model is more overfitted to private training data.
Second, when the recovered image $\tilde x_i^j$ is close to the training data, $F_c(\tilde x_i^j)$ should be \emph{consistently} higher on the correct label because training data are often more resistant to transformations than non-training data~\cite{choquette2021label}. Based on these intuitions, we use the discrimination metric as the foundation of the pseudo label predictor in \method.

\noindent\textbf{Pseudo Label Predictor.}\quad
Given the discrimination metric $\cal M$, we next describe in detail how we leverage $\cal M$ to infer the pseudo prediction label considering different sampled latent vectors, which aims to approximate the ground truth.
% is inferred from the auxiliary information with our discrimination metric.
We first sample a set of $n$ latent vectors randomly and compute $\mathcal{M}$ for all class labels.
The pseudo label predictor chooses the label with the maximum discrimination metric score as the predicted label $\hat c$:
% In previous works where ground truth label is available, we can directly use the ground truth as the target for optimization.
% However in our setting, the ground truth may not be accessible.
% Our label predictor infers predicted label by leveraging the auxiliary information and target model.
\vspace{-0.5em}
\begin{equation}
    \hat c = \mathop{\arg\max}_{c\in [1,C]}\mathcal{M}(c)
    \label{eq:base}
\end{equation}

\vspace{-0.5em}
We defer the detailed algorithm for label prediction with $\mathcal{M}$ to the appendix. Note that there are various design choices for the discrimination metric $\mathcal{M}$, 
% \sout{\eg the resulting $\mathcal{M}$ after removing the transformations in Equation~\ref{eq:m2}}.
\eg, the average confidence on only the recovered images without including their transformed versions.
% \fan{hmm, this so-called variant in the current formulation is actually equivalent to the current $\cal M$ with $m=0$.}
It is clear that more advanced $\mathcal{M}$ will provide more accurate pseudo label predictors.
We will analyze the performance of the pseudo label predictor given different designs of $\mathcal{M}$ in Section~\ref{sec:pred_acc}.

% \fan{logic: in previous works where groudtruth is available, we can directly use; here gt label is unavailable, so we need to use other information...}

% \fan{defer the analysis about label exposure to experiments; present $\mathcal M_2$ in main paper and say there are other metrics ...}

\vspace{-1em}
\subsection{Latent Vector Selector}
\vspace{-0.5em}
\label{sec:search}

In addition to the availability of ground truth labels, another challenge during private data recovery is that we may not have $whitebox$ access to the target model.
In systems where machine learning is used as a service (MLaaS), the adversary can only query the model and the prediction vector is returned from the service provider.
All internal computations and model parameters are unknown to the adversary.
In previous work \cite{zhang2020secret}, the adversary can directly optimize the latent vector $z$ to maximize the target model's confidence given a known ground truth label, which is less practical.
Without the whitebox access, performing back-propagation with the target model is infeasible in our practical case.
% \fan{this paragraph is confusing to me.. this step of latent selector is comparable to the secret revealer's random selection, and our optimization step is comparable to their optimization. Why do we compare our latent selector with their optimization?}

To tackle this problem, we design a \textit{latent vector selector} to first randomly sample $n$ random latent vectors, and then select the ones which lead to their recovered data classified as the predicted label from the pseudo label predictor.
% To tackle this problem, we design a \textit{latent vector selector} to first choose an initial latent vector from a set of $n$ random latent vectors, and then select the ones iteratively which lead to the data recovery with the highest confidence for the predicted label by the label predictor.
% \fan{If we never assume we can access ground truth label, shall we mention it here? maybe we can have a separate short paragraph after we finish introducing the entire pipeline to discuss how our framework can also leverage the known ground truth label.}
If there is no latent vector that leads to the recovered images which can be classified as the predicted label consistently, the selector returns a randomly sampled latent vector from the prior distribution. Otherwise, it returns the latent vector which has the highest confidence of the predicted label. 
We omit the detailed algorithm to the appendix. 
% about how the we select the initial latent vector with the predicted labels consistently.

% \begin{algorithm}
% \caption{Latent Vector Selection}
% % \fan{similar comments as for Algorithm 1} \bl{maybe leave this one here}
% \label{alg:select}
% \begin{algorithmic}
% \small
% \State \textbf{Input:} $x_{ns}$: corrupted private data, $F$: target model, $n$: number of randomly sampled latent vectors, $y$: predicted label or ground truth label

% \State $z_{bank} \gets $ sample $n$ latent vectors from $\mathcal{N}(0, 1)$
% \State $z_{bank\_y} \gets $ $\{z \in z_{bank} | \arg\max_{c\in[1,C]} F_c(G(z, x_{ns}))=y\} $
% % iterate over $z_{bank}$, select $z \in z_{bank}$ where $\arg\max_{c\in[1,C]} F_c(G(z, x_{ns}))=y$
% \If{$z_{bank\_y}$ is not empty}
%     % \State \sout{$c_{y} \gets F_y(G(z_{bank\_y}, x_{ns}))$ for $z \in z_{bank\_y}$}
%     % \State \sout{$\mathit{index} \gets \arg\max c_{y}$}
%     % \State \sout{$\hat z \gets c_{y}[\mathit{index}]$}
%     \State $\hat z \gets \arg\max_{z \in z_{bank\_y}} F_y(G(z, x_{ns}))$
% \Else
%     % \State \sout{$z \sim \mathcal{N}(0, 1)$}
%     % \State \sout{$\hat z \gets z$}
%     \State $\hat z \sim \mathcal{N}(0, 1)$
% \EndIf

% \State \textbf{return} $\hat z$

% \end{algorithmic}
% \end{algorithm}

\vspace{-1em}
\subsection{\method Optimization}
\vspace{-0.5em}
\label{sec:opt}
To put every proposed component within \method together, we perform joint optimization to maximize the consistent label prediction likelihood of recovered images indicated by the discrimination metric (\ie, identity loss),  while keeping the recovered images realistic (\ie, discriminator loss).
In the \textit{whitebox} setting, we perform backpropagation on the target model with identity loss $\mathcal{L}_{\rm id}$.
$\mathcal{L}_{\rm id}$ encourages the generated images to achieve consistently high label prediction likelihood given the target model for class label $c$.
\begin{equation}
    \mathcal{L}_\text{id} = - \log[F_c(G(z, x_{ns}))]
\end{equation}

We utilize discriminator loss as regularization to penalize unrealistic images.
\begin{equation}
    \mathcal{L}_{\rm disc} = - D(G(z, x_{ns}))
\end{equation}

Then we initialize $z$ with $\hat z$ returned by our latent vector selector and optimize $z$ with the following objective function:
\begin{equation}
    \label{eq:opt-main}
    \hat{z}_{\rm whitebox}^* = \mathop{\arg\min}_z \mathcal{L}_{\rm disc} + \lambda_{\rm id} \mathcal{L}_{\rm id}
\end{equation}

In the \textit{blackbox} setting,  we perform the latent vector selection optimization only with the discriminator loss since the target model is not locally accessible:
\begin{equation}
    \hat{z}_{\rm blackbox}^* = \mathop{\arg\min}_z \mathcal{L}_{\rm disc}
\end{equation}
Note that in the blackbox setting where we have the ground truth labels, the identity loss is still minimized by the latent vector selector through random sampling based on the prediction vector of the target model, guaranteeing that the recovered image is close to the density region of the ground truth identity.

\vspace{-0.5em}
\subsection{Discussion}
\vspace{-0.5em}
Our proposed \method works under a wide range of scenarios regarding different types of prior knowledge.
See Table~\ref{tab:scenario} for the scenarios under which \method and existing methods can be applied.
Although EMI theoretically works in $blackbox$ cases with ground truth labels, its performance and efficiency dramatically suffer against deep models.
Although \method still requires non-sensitive private data as prior knowledge, such assumption is realistic as image corruption is often leveraged for privacy protection by individuals \cite{zhang2020secret}.
Furthermore, if the knowledge of ground truth labels is available, it can be incorporated into \method conveniently. 
More details are deferred to the appendix.

% ~\yhc{Can we add an additional column ``effective against deep models'' in the table?}\bl{good idea, let's add it} 
In conclusion, \method is more practical without requiring whitebox access to the target model or the ground truth label. 
In addition, \method is very efficient and applicable to high-dimensional image data considering deep models as the target model as shown in our evaluation (Section~\ref{sec:exp}).

\begin{table}[ht]
    \centering
    \caption{{\small Comparison with existing methods on the information required by the adversary to recover private training data.
    The symbol \cxmark\space means that in theory, the method can work without the information, but the actual performance on deep models is bad.}}
    %  \yhc{ Why are methods listed in this table different from the ones in evaluation? Seems that PII is in evaluation but not in this table, EMI is in this table but not in evaluation} \bl{indeed, change EMI to PII here?}}
    \resizebox{0.75\linewidth}{!}{
    \renewcommand{\arraystretch}{1.2}
    \begin{tabular}{|c|c|c|c|}
    \hline
    \textbf{Methods} & \textbf{Non-sensitive Data?} & \textbf{Whitebox Access?} & \textbf{Ground Truth Label?} \\
    \hline
    PII~\cite{yang2019diversity} & \cmark & \xmark & \xmark \\
    EMI~\cite{fredrikson2015model} & \xmark & \cxmark & \cmark \\
    GMI~\cite{zhang2020secret}     & \cxmark & \cmark & \cmark \\
    \method & \cmark & \xmark & \xmark \\
    \hline
    \end{tabular}
    }
    \label{tab:scenario}
    \vspace{-2em}
\end{table}
 
\section{Experiments}
\label{sec:exp}
In this section, we first present the experimental setup. Then, we introduce the evaluation protocols and report the attack performance respectively. We also evaluate the robustness of \method against the purification defense \cite{yang2020defending}. In the end, we describe some ablation studies to better understand our method. 

\subsection{Experimental Setup}
% \vspace{-1em}
\label{sec:setup}
\noindent\textbf{Datasets.}\quad
We evaluate \method on two face datasets: (1) CelebA \cite{liu2015faceattributes} which contains 202,599 face images of 10,177 identities.
We filter out those identities with 25 or fewer images and randomly select 25,000 images of 1,000 identities as private data $D_{\rm pri}$.
We also randomly select 50,000 images of 2,000 identities from the rest as adversary's public data $D_{\rm pub}$.
There exist no overlapped identities between $D_{\rm pub}$ and $D_{\rm pri}$.
(2) FaceScrub \cite{ng2014data} which consists of 106,863 face images of 530 identities. We use the images of 250 identities as $D_{\rm pri}$ and images of another 250 identities as $D_{\rm pub}$.
We further split $D_{\rm pri}$ into $D_{\rm pri}^{\rm train}$ and $D_{\rm pri}^{\rm test}$ for training and testing.
All the images are cropped and resized to $64 \times 64$.

\noindent\textbf{Prior Information.}\quad
We consider two types of prior information that the adversary has access to: corrupted images by center mask and face T mask following the standard setting in~\cite{zhang2020secret}.
Center mask blocks the center part of the private image, but the mouth information may still be exposed.
Face T mask completely hides the identity revealing features of the face image.

\noindent\textbf{Model Architectures.}\quad
We perform evaluation on \textit{target models} with various architectures:
(1) \texttt{VGG16} \cite{simonyan2014very};
(2) \texttt{ResNet152} \cite{he2016deep};
(3) \texttt{face.evoLVe} \cite{cheng2017know} with an \texttt{IR50} backbone;
(4) \texttt{ViT-B\_16} \cite{dosovitskiy2020image} 
% which is a \texttt{ViT-Base} model with patch resolution of $16 \times 16$.
We utilize \texttt{IR152} \cite{cheng2017know} as the \textit{evaluation model} to predict the identity of input images.
Both \texttt{VGG16} and \texttt{ResNet152} are pre-trained on \texttt{ImageNet}~\cite{deng2009imagenet}.
\texttt{face.evoLVe} and the evaluation model are pre-trained on \texttt{MS-Celeb-1M} \cite{guo2016ms}.
\texttt{ViT-B\_16} is pre-trained on \texttt{ImageNet21k}~\cite{deng2009imagenet}.
% We resize the images to fit the input size of different model architectures.
The architecture of \method generation backbone is adopted from~\cite{zhang2020secret}.

% \fan{details of $m$ transformations, $n$ in label predictor, and $n$ in latent vector selector. btw, do we use two different $n$ for these two stages? maybe we can call them $n_1$ and $n_2$?}https://www.overleaf.com/project/61897512989f5fdcdc5f25fc

\noindent\textbf{Baselines.}\quad
We compare \method with the \textit{state-of-the-art} model inversion attack GMI \cite{zhang2020secret}.
GMI assumes the adversary has access to the ground truth labels and performs optimization with identity loss and discriminator loss.
We also compare our results with pure image inpainting (PII)~\cite{yang2019diversity}, which only optimizes the discriminator loss for generating realistic images.
Latent vectors of both GMI and PII are sampled randomly from Gaussian distribution. 
We do not compare with EMI \cite{fredrikson2015model}, since it has been demonstrated in \cite{zhang2020secret} that the effectiveness of EMI is quite limited against deep models. We defer additional details regarding model training and attack to the appendix.

\subsection{Evaluation Protocols}

We consider two principles for evaluating the privacy attack performance: ``how much privacy sensitive identity information can be recovered'' and ``how well the recovered data can perform in downstream tasks''.

% \noindent\textbf{Baselines.}\quad
% We compare \method with the \textit{state-of-the-art} model inversion attack GMI \cite{zhang2020secret}.
% GMI assumes the adversary has access to the ground truth predictions and performs optimization with identity loss and discriminator loss.
% We also compare our results with pure image inpainting (PII)~\cite{yang2019diversity}, which only optimizes the discriminator loss for generating realistic images.
% Latent vectors of both GMI and PII are sampled randomly from Gaussian distribution. 

Corresponding to the two principles, we evaluate the privacy attack performance by \textit{attack accuracy} under the following two protocols:
\begin{itemize}
    \item Protocol 1: Train the evaluation model on the private data, and evaluate on the recovered data.
    \item Protocol 2: Train the evaluation model on the recovered data, and evaluate on the private data.
\end{itemize}

Protocol 1 was introduced in \cite{zhang2020secret}, which evaluates \textit{instance-level} privacy recovery.
However, we demonstrate that even if some instances are not recovered correctly, the recovered data can be used 
for downstream tasks, \eg, training another classification model.
The adversary can potentially use the trained evaluation model for malicious purposes, \eg, performing unauthorized face recognition on private identities with significantly higher accuracy than the target model itself.
Thus, we propose  Protocol 2, which aims to evaluate \textit{distribution-level} privacy recovery.
In addition, a common goal of the adversary to reconstruct the private data is to leverage such data for other downstream tasks, and therefore Protocol 2 explicitly reflects the utility of the recovered data.

For Protocol 1, we train the evaluation model on $D_{\rm pri}^{\rm train}$ and the resulting evaluation model achieves 98.0\% classification accuracy over the private identities on $D_{\rm pri}^{\rm test}$. 
For Protocol 2, we first perform the attack on all corrupted private images $D_{\rm pri}$---for each corrupted image $x_{ns} \in D_{\rm pri}$, we recover an image $\hat x = G(\hat z^*, x_{ns})$ via \method, with label $\hat c=\arg\max_{c\in[1,C]} F_c(\hat x)$.
% \sout{We then construct a recovered private set $D_{\rm rec}$ with image-label pairs: $(\hat x, F(\hat x))$. }\fan{$F$ does not denote the label}
We then compose the recovered images into a recovered private set $D_{\rm rec}$, which is separated into $D_{\rm rec}^{\rm train}$ and $D_{\rm rec}^{\rm valid}$ by 4:1.
We train the evaluation model on $D_{\rm rec}^{\rm train}$ with $D_{\rm rec}^{\rm valid}$ as the validation set.
We then evaluate the model performance on $D_{\rm pri}^{\rm test}$. 
% \fan{why not evaluate on the entire private dataset $D_{\rm pri}$?}

% ~\yhc{Need to add some intuitive explanations on why we want to add protocol 2 to the evaluation}

We also report Peak Signal-to-Noise Ratio (PSNR) \cite{hore2010image} between original and recovered private data, which reflects the \textit{pixel-level} reconstruction quality of our attack.
Note that the recovered data can still reveal identity information even if the generated image is not close to the ground truth image pixel-wise.
For example, the recovered images can exhibit variations in pose and light condition while keeping the identity.

\subsection{Attack Performance}
\noindent\textbf{Whitebox Attacks.}\quad
Table~\ref{tab:whitebox} compares the performance of \method with baseline methods on CelebA.
See the appendix for results on FaceScrub.
% \fan{Do we have a same full set of results for both datasets? I do not see a similar pointer sentence for next two parts... btw, now we have two datasets, it may be better to emphasize the dataset in the table/figure caption.}
% \zw{Under the current plan, the exps for FaceScrub is only for VGG16, whitebox setting. Maybe we can run more experiments if time permits :)}

\begin{table}[ht]
\centering
\caption{\textit{Whitebox} attack performance on CelebA. See the Ground Truth Label column for whether ground truth label is provided for each attack method.}
 
\label{tab:whitebox}
\resizebox{\linewidth}{!}{
\begin{tabular}{ccccccccc}
\toprule
\multirow{2}{*}{\textbf{Target Model}}       & \multirow{2}{*}{\textbf{Methods}} & \multirow{2}{*}{\textbf{\begin{tabular}[c]{@{}c@{}}Ground Truth\\ Label\end{tabular}}} & \multicolumn{3}{c}{\textbf{Center Mask}}                    & \multicolumn{3}{c}{\textbf{Face T Mask}}                    \\ \cmidrule{4-6} \cmidrule{7-9}
                                      &                                   &                                                                                        & \textbf{Protocol 1} & \textbf{Protocol 2} & \textbf{PSNR}   & \textbf{Protocol 1} & \textbf{Protocol 2} & \textbf{PSNR}   \\ \hline
\multirow{4}{*}{VGG16}       & PII                      & \xmark                                                                                       & 0.423               & 0.561               & 27.583          & 0.166               & 0.363               & 26.276          \\
                                      & GMI                      & \cmark                                                                                       & 0.569               & 0.955               & 27.587          & 0.305               & 0.928               & 26.240          \\
                                      & SecretGen                & \xmark                                                                                       & 0.584               & 0.928               & 27.955          & 0.312               & 0.793               & 26.632          \\
                                      & SecretGen                & \cmark                                                                                       & \textbf{0.639}      & \textbf{0.965}      & \textbf{28.071} & \textbf{0.377}      & \textbf{0.931}               & \textbf{26.821} \\ \hline
\multirow{4}{*}{ResNet152}   & PII                      & \xmark                                                                                       & 0.403               & 0.719               & 26.892          & 0.170               & 0.555               & 26.117          \\
                                      & GMI                      & \cmark                                                                                       & 0.556               & 0.965               & 27.177          & 0.295               & \textbf{0.946}               & 26.482          \\
                                      & SecretGen                & \xmark                                                                                       & 0.595               & 0.948               & 27.506          & 0.324               & 0.884               & 26.821          \\
                                      & SecretGen                & \cmark                                                                                       & \textbf{0.618}      & \textbf{0.971}      & \textbf{27.587} & \textbf{0.349}      & 0.945               & \textbf{26.967} \\ \hline
\multirow{4}{*}{face.evoLVe} & PII                      & \xmark                                                                                       & 0.267               & 0.455               & 27.317          & 0.122               & 0.343               & 26.356          \\
                                      & GMI                      & \cmark                                                                                       & 0.595               & 0.946               & 27.444          & 0.467               & 0.935               & 26.563          \\
                                      & SecretGen                & \xmark                                                                                       & 0.551               & 0.841               & 27.613          & 0.274               & 0.630               & 26.562          \\
                                      & SecretGen                & \cmark                                                                                       & \textbf{0.788}      & \textbf{0.963}      & \textbf{27.781} & \textbf{0.695}      & \textbf{0.954}      & \textbf{26.827} \\ \hline
\multirow{4}{*}{ViT} & PII                      & \xmark                                                                                       & 0.380               & 0.389               & 26.698          & 0.173               & 0.306               & 26.377          \\
                                      & GMI                      & \cmark                                                                                       & 0.482               & 0.893               & 24.907          & 0.214               & 0.715               & 24.624          \\
                                      & SecretGen                & \xmark                                                                                       & 0.451               & 0.634               & \textbf{26.811}          & 0.246               & 0.528               & 26.471          \\
                                      & SecretGen                & \cmark                                                                                       & \textbf{0.551}      &   \textbf{0.950}    & 26.607 & \textbf{0.326}      & \textbf{0.913}      & \textbf{26.609} \\
\bottomrule
\end{tabular}
\vspace{-2em}
}
\end{table}

We can see that \method significantly outperforms GMI under both Protocol 1 and Protocol 2 if the ground truth label is given.
Without such information, with the proposed pipeline especially the pseudo label predictor, \method still achieves comparable performance with GMI under Protocol 1.
Under Protocol 2, GMI with ground truth label performs better than \method without ground truth label.
The reason is that if the predicted pseudo label is incorrect, our pseudo label predictor and optimization push the recovery to be closer to the wrong identity.
However, we still outperform PII by a large margin.

We also observe that attack accuracy under Protocol 2 is much higher than that under Protocol 1. The reason is that Protocol 1 and 2 work at different levels: Protocol 1 evaluates how much ``detailed" information the recovered images contain, while Protocol 2 evaluates how much distributional information we can recover by training another model based on the reconstructed data. Clearly, Protocol 2 is relatively easier by recovering distributional level information and thus achieves higher scores.
We believe such observations will inspire interesting future work and narrow down such a gap.

\noindent\textbf{Blackbox Attacks.}\quad
In the $blackbox$ setting, the adversary is not capable of performing backpropagation with the target model.
We make the following changes to our attack pipeline: 
(1) In Section \ref{sec:gen-backbone}, when training the generation backbone, we use a public feature extractor from \cite{cheng2017know} pre-trained on \texttt{MS-Celeb-1M} to substitute the target model for extracting the feature embeddings in computing the diversity loss ($\mathcal{L}_{\rm div}$, Eqn.~\eqref{eq:div}); 
(2) In Section \ref{sec:opt}, when performing \method optimization, we remove the identity loss $\mathcal{L}_{\rm id}$ and optimize the selected latent vector only with discriminator loss $\mathcal{L}_{\rm disc}$.

Table~\ref{tab:blackbox} compares our results with PII under the \textit{blackbox} setting on CelebA.
The only difference for PII under $blackbox$ and $whitebox$ scenarios is whether the target model is accessed when training the generation backbone.
We can see that with the ground truth labels, \method significantly outperforms PII.
Without ground truth labels, which is the most general case, we still outperform PII by a large margin.
As far as we are concerned, we are the first to propose an effective model inversion attack against deep classification models under the \textit{blackbox} case without ground truth label.

We note that GMI (with ground truth label) performs better on \texttt{face.evoLVe} than SecretGen (without ground truth label), as shown in Table~\ref{tab:whitebox} and Table~\ref{tab:blackbox}. Under this setting, attack performance is largely dependent on the pseudo label predictor. We demonstrate that the label prediction accuracy of \texttt{face.evoLVe} is significantly lower than that of \texttt{VGG16} and \texttt{ResNet152} in the appendix. We believe the reason is that \texttt{face.evoLVe} is less overfitted due to the difference in pre-training datasets. (\texttt{face.evoLVe} is pre-trained on \texttt{MS-Celeb-1M} while others are on \texttt{ImageNet}.)

% Note that although EMI \cite{fredrikson2015model} also supports $blackbox$ attacks if the ground truth prediction is given, their $whitebox$ performance is significantly lower than PII as demonstrated in \cite{zhang2020secret}.
% Besides, there exist no efficient implementation of their algorithm against deep models.

\begin{table}[t]
\centering
\caption{\small \textit{Blackbox} attack performance on CelebA. We report results for both cases where the adversary has or does not have ground truth labels. (Note: GMI does not support blackbox attack, and PII in the blackbox setting does not use the target model.) 
% PII does not queries the target model in the $blackbox$ setting.
}\label{tab:blackbox}
\resizebox{\linewidth}{!}{
\begin{tabular}{ccccccccc}
\toprule
\multirow{2}{*}{\textbf{Methods}} & \multirow{2}{*}{\textbf{Target Model}} & \multirow{2}{*}{\textbf{\begin{tabular}[c]{@{}c@{}}Ground Truth\\ Label\end{tabular}}} & \multicolumn{3}{c}{\textbf{Center Mask}} & \multicolumn{3}{c}{\textbf{Face T Mask}} \\ \cline{4-9} 
 &  &  & \textbf{Protocol 1} & \textbf{Protocol 2} & \textbf{PSNR} & \textbf{Protocol 1} & \textbf{Protocol 2} & \textbf{PSNR} \\
 \midrule
PII & Any & \xmark & 0.216 & 0.759 & 27.319 & 0.081 & 0.484 & 25.705 \\ \cline{1-9} 
\multirow{8}{*}{SecretGen} & \multirow{2}{*}{VGG16} & \xmark & 0.351 & 0.915 & 27.638 & 0.164 & 0.837 & 26.045 \\
 &  & \cmark & \textbf{0.380} & \textbf{0.955} & \textbf{27.737} & \textbf{0.377} & \textbf{0.927} & \textbf{26.821} \\ \cline{2-9} 
 & \multirow{2}{*}{ResNet152} & \xmark & 0.334 & 0.933 & 27.737 & 0.152 & 0.765 & 26.144 \\
 &  & \cmark & \textbf{0.347} & \textbf{0.959} & \textbf{27.840} & \textbf{0.172} & \textbf{0.886} & \textbf{26.284} \\ \cline{2-9} 
 & \multirow{2}{*}{face.evoLVe} & \xmark & 0.447 & 0.711 & 27.568 & 0.156 & 0.353 & 25.787 \\
 &  & \cmark & \textbf{0.603} & \textbf{0.894} & \textbf{27.694} & \textbf{0.305} & \textbf{0.586} & \textbf{26.002} \\ \cline{2-9} 
 & \multirow{2}{*}{ViT} & \xmark & 0.285 & 0.709 & 27.480 & 0.119 & 0.685 & 25.828 \\
 &  & \cmark & \textbf{0.335} & \textbf{0.924} & \textbf{27.665} & \textbf{0.160} & \textbf{0.902} & \textbf{26.123} \\
 \bottomrule
\end{tabular}

}

\end{table}

\begin{figure}[t]
    \centering
    \includegraphics[width=0.75\textwidth]{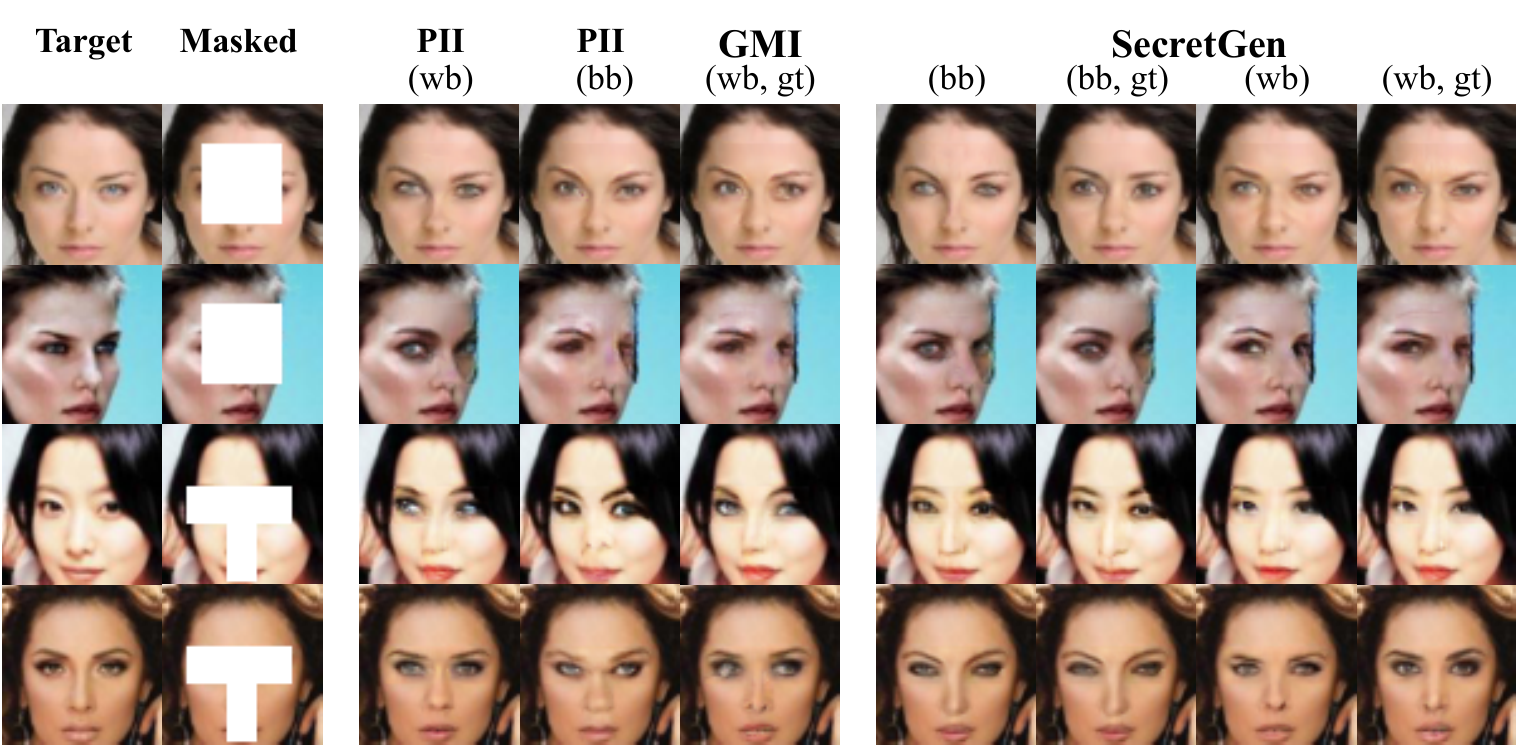}
    \caption{Qualitative results of \method on CelebA. ``bb''/``wb'' indicates the method requires \textit{blackbox}/\textit{whitebox} access to the model. ``gt'' indicates the method requires ground truth labels.}
    \label{fig:qualitative}
    \vspace{-2em}
\end{figure}

\noindent\textbf{Qualitative Results.}\quad
In Fig.~\ref{fig:qualitative} we exhibit the images recovered with \method on CelebA to demonstrate that our recovered images are both identity-revealing and visually plausible.
We also show qualitative results of PII and GMI for comparison.
From the figure, we see that although all of the three methods generate realistic images, PII cannot effectively recover the original identity of private data, while \method is more effective in identity revealing. More examples are shown in the appendix.

\vspace{-1em}
\subsection{Robustness Evaluation}
We evaluate the robustness of our proposed method against purification defense~\cite{yang2020defending}, which has been shown to effectively defend against model inversion attacks while inducing negligible utility loss.
We use Purifier \Romannum{1} in \cite{yang2020defending} which is specialized for model inversion attacks.
We follow the default architectures and settings for training the purifier.
See Table~\ref{tab:robust} for quantitative results on CelebA against VGG16 under the blackbox setting.
We also assume the ground truth label is not provided.
We do not evaluate the whitebox setting because the adversary can simply remove the purifier and directly attack the original private model.
It can be seen that attack accuracy slightly decreases after the defense, but still outperforms the baseline by a large margin.
Therefore, our method is robust against \cite{yang2020defending}.
% since \method does not require a precise model output.

\begin{table}[bt]
    \centering
    \caption{Robustness evaluation for \method against prediction purification on CelebA. Target model: \texttt{VGG16}. Blackbox setting.}
    \resizebox{0.8\textwidth}{!}{
        \begin{tabular}{ccccccc}
        \toprule
        \multirow{2}{*}{\textbf{Methods}} & \multicolumn{3}{c}{\textbf{Center Mask}} & \multicolumn{3}{c}{\textbf{Face T Mask}} \\ \cline{2-7} 
         & \textbf{Protocol 1} & \textbf{Protocol 2} & \textbf{PSNR} & \textbf{Protocol 1} & \textbf{Protocol 2} & \textbf{PSNR} \\
        \midrule
        PII & 0.216 & 0.759 & 27.319 & 0.081 & 0.484 & 25.705 \\
        SecretGen & 0.351 & 0.915 & 27.638 & 0.164 & 0.837 & 26.045 \\
        SecretGen (purified) & 0.328 & 0.913 & 27.590 & 0.151 & 0.747 & 26.007 \\
        \bottomrule
        \end{tabular}
    }
    \label{tab:robust}
    \vspace{-1em}
\end{table}

\subsection{Ablation Studies}
\label{sec:pred_acc}

\noindent\textbf{Discrimination Metrics.}\quad
As discussed in Section~\ref{sec:disc-metric}, there may exist various choices for the discrimination metric.
One intuitive choice may be derived by removing the transformations from our current discrimination metric $\cal M$ (Eqn.~\eqref{eq:metric}), and the simplified discrimination metric is defined as follows:
\begin{equation}
\label{eq:m2}
    \mathcal{M}'(c;n) \triangleq \frac{1}{n}\sum_{i=1}^nF_c(\hat x_i),\quad \forall c \in [1,C].
\end{equation}

We perform an \textit{end-to-end} ablation study on \texttt{face.evoLVe} and CelebA.
We remove the transformations in our pseudo label predictor and substitute $\mathcal{M}$ with $\mathcal{M'}$.
Quantitative results on \texttt{face.evoLVe} are shown in Table~\ref{tab:ablation}. 
See the appendix for results regarding other model architectures.
We conclude that incorporating transformations improves the performance of our framework for most model architectures that we used for evaluation.

\begin{table}[ht]
\centering
% \vspace{-1em}
\caption{Attack accuracy of \method with and without transformations on CelebA. Evaluated on 3,200 private instances under Protocol 1. Target model: \texttt{face.evoLVe}.}\label{tab:ablation}
\resizebox{0.65\linewidth}{!}{
\begin{tabular}{ccccc}
\toprule
\multirow{2}{*}{\textbf{Metric}} & \multicolumn{2}{c}{\textbf{Center Mask}} & \multicolumn{2}{c}{\textbf{Face T Mask}} \\ \cline{2-5} 
                                  & \textbf{Attack Acc}   & \textbf{PSNR}    & \textbf{Attack Acc}   & \textbf{PSNR}    \\ \hline
w/o transformation                & 0.528                 & 27.505           & 0.256                 & 26.527           \\
w/ transformation                 & \textbf{0.550}        & \textbf{27.522}  & \textbf{0.273}        & 26.527 \\
\bottomrule
\end{tabular}
}
\vspace{-1em}
\end{table}

\begin{figure}[!htb]
\begin{minipage}{0.48\textwidth}

\newlength{\utilheightreward}
\settoheight{\utilheightreward}{\includegraphics[width=.75\linewidth]{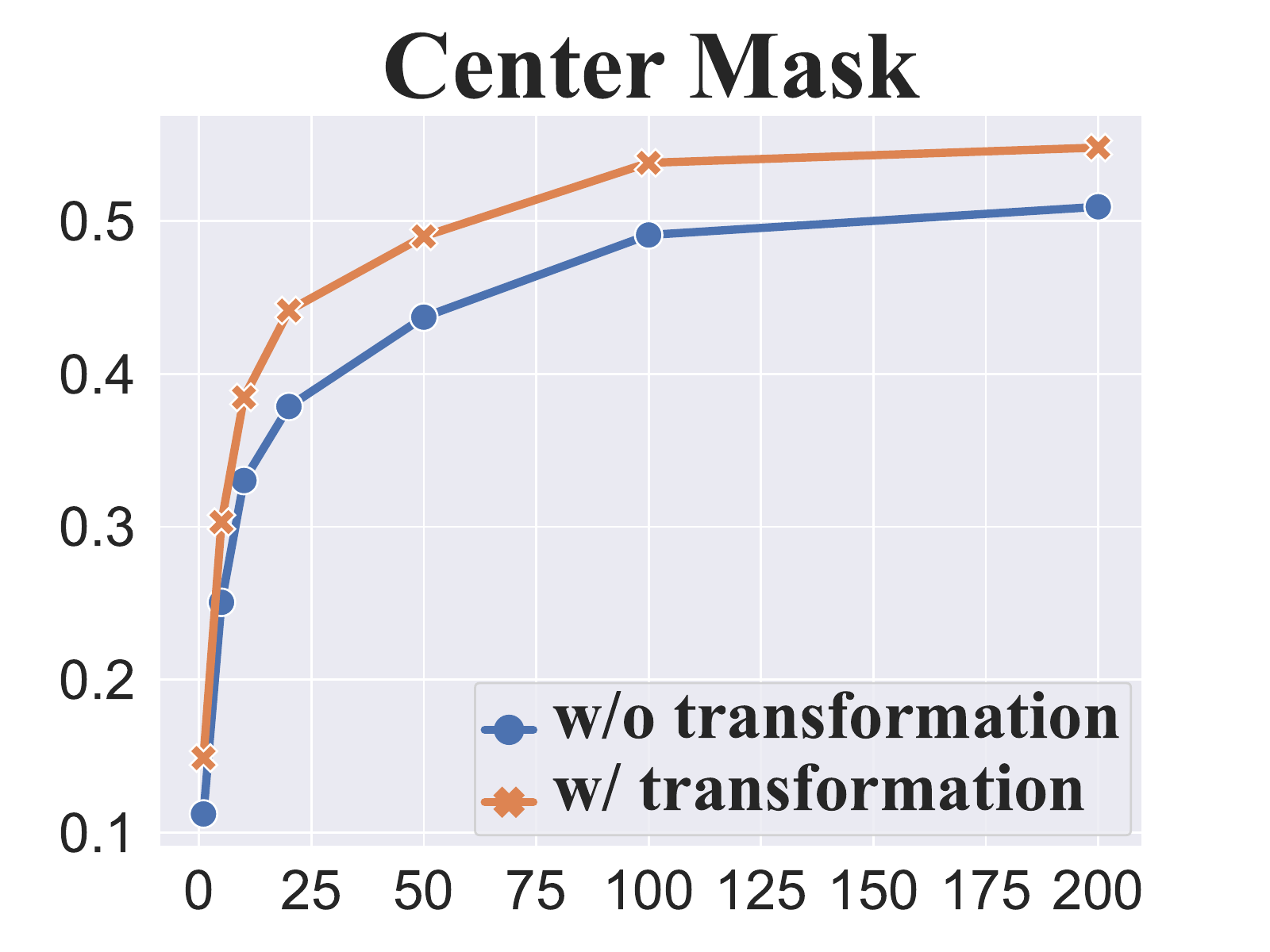}}%

\newlength{\legendheight}
\setlength{\legendheight}{0.27\utilheightreward}%

\newcommand{\rowname}[1]% #1 = text
{\rotatebox{90}{\makebox[\utilheightreward][c]{\tiny #1}}}

\centering

{
\renewcommand{\tabcolsep}{10pt}

\begin{subtable}
\centering
\resizebox{\linewidth}{!}{%
\begin{tabular}{@{}p{3mm}@{}c@{}c@{}}
        % & \makecell{\small{\textbf{VGG16}}}
\rowname{\small Accuracy }&
\includegraphics[height=\utilheightreward]{figs/label_ir50_center.pdf}& 
\includegraphics[height=\utilheightreward]{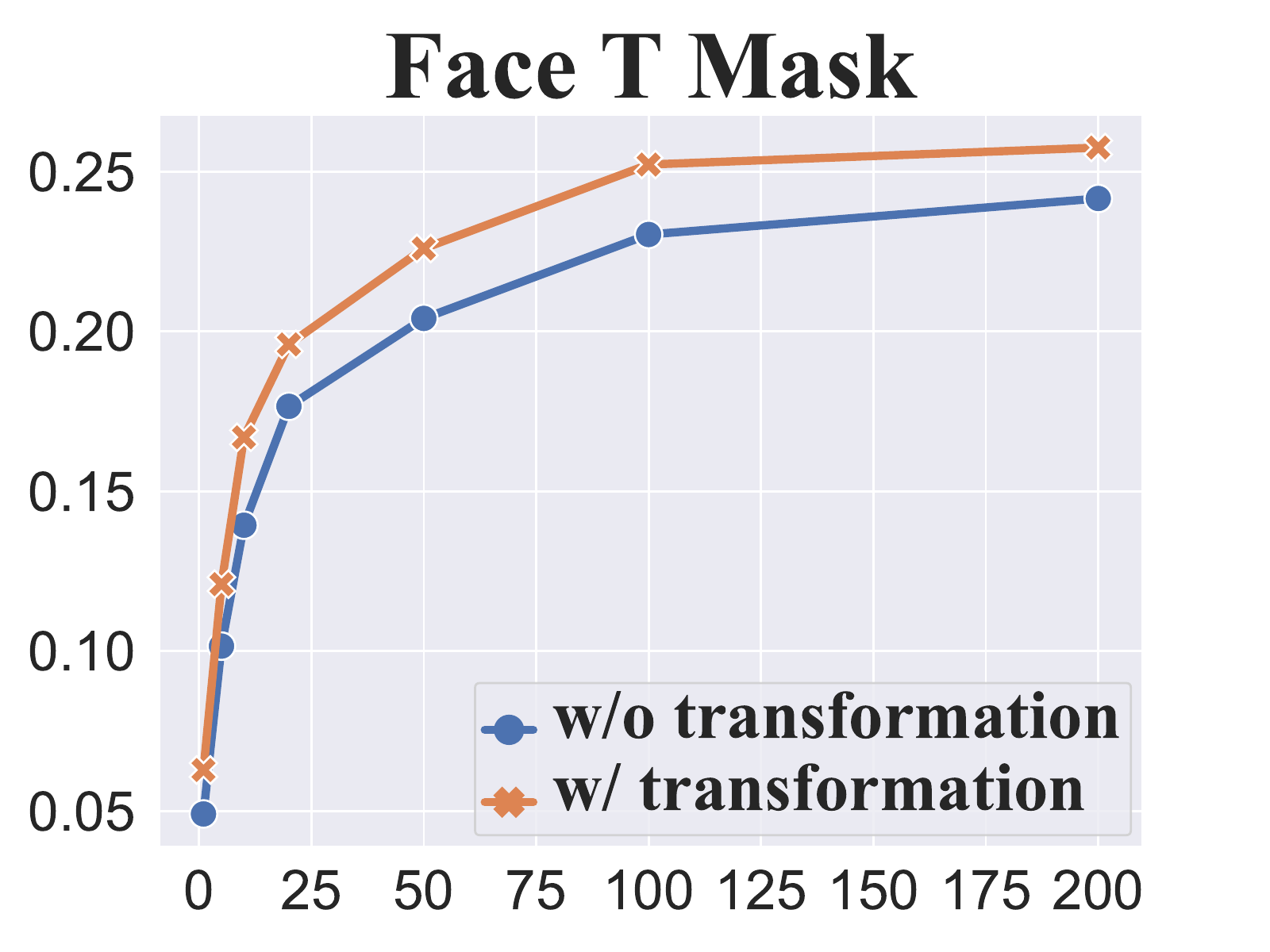}
\\[-1.5ex]
        & \makecell{\small{$n$}}
        & \makecell{\small{$n$}}
\end{tabular}
}
\end{subtable}

}

\caption{\small Label prediction accuracy with and without transformations on CelebA. We plot the label prediction accuracy w.r.t. the number of random latent vectors~$n$. Target model: \texttt{face.evoLVe}.
}
\label{fig:label_prediction_acc}

\end{minipage}
\hfill
\begin{minipage}{0.48\textwidth}
\newlength{\utilheighttrans}
\settoheight{\utilheighttrans}{\includegraphics[width=.75\linewidth]{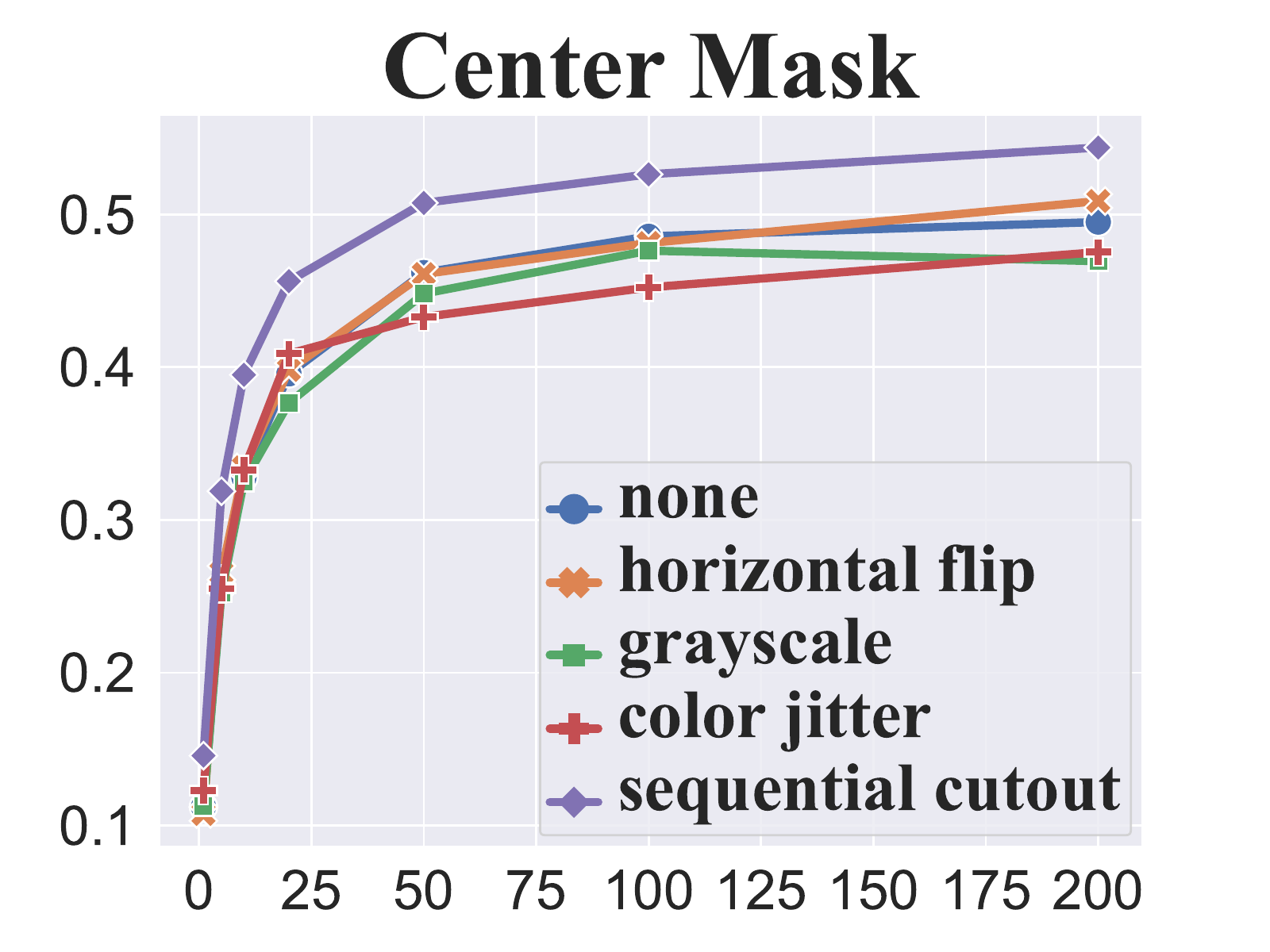}}%

\newcommand{\rowname}[1]% #1 = text
{\rotatebox{90}{\makebox[\utilheighttrans][c]{\tiny #1}}}

\centering

{
\renewcommand{\tabcolsep}{10pt}

\begin{subtable}
\centering
\resizebox{\linewidth}{!}{%
\begin{tabular}{@{}p{3mm}@{}c@{}c@{}}
        % & \makecell{\small{\textbf{VGG16}}}
\rowname{\small Accuracy }&
\includegraphics[height=\utilheighttrans]{figs/ir50_center.pdf}& 
\includegraphics[height=\utilheighttrans]{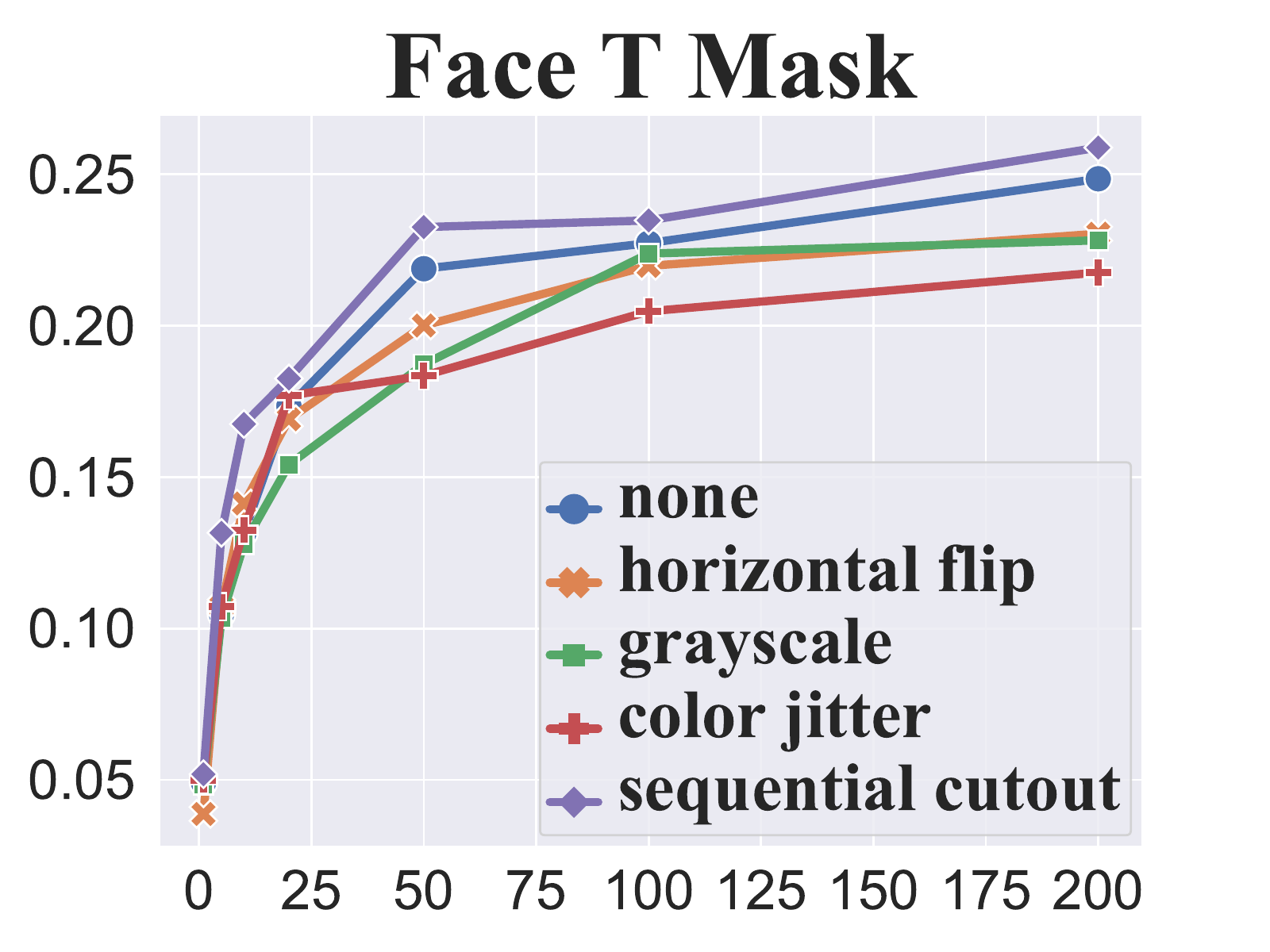}
\\[-1.5ex]
        & \makecell{\small{$n$}}
        & \makecell{\small{$n$}}
\end{tabular}
}
\end{subtable}

}

\caption{\small Label prediction accuracy with different transformations on CelebA. We plot the label prediction accuracy w.r.t. the number of randomly sampled latent vectors $n$. Target model: \texttt{face.evoLVe}.}

\label{fig:various_trans}
\end{minipage}

\vspace{-1em}
\end{figure}

To further understand why and how transformations help, we compare the performance of pseudo label predictor equipped with $\mathcal{M}$ and $\mathcal{M}'$. We evaluate the performance of pseudo label predictor using \textit{label prediction accuracy}, which measures the percentage of the predicted labels matching the ground truth labels.
We plot out the label prediction accuracy with $\mathcal{M}$ and $\mathcal{M}'$ on 3,200 recovered images for \texttt{face.evoLVe} with varying $n$ in Fig.~\ref{fig:label_prediction_acc}.
We observe that our pseudo label predictor can predict the pseudo labels more accurately if transformations are incorporated.
See the appendix for results of other model architectures.

\renewcommand{\thesubfigure}{\alph{subfigure}}
\newcommand{\mycaption}[1]% #1 = caption
{\refstepcounter{subfigure}\textbf{(\thesubfigure) }{\ignorespaces #1}}

% \fan{Put Fig 4 in the same row with Fig 5. Align the y-axis. If needed, can discuss in the main text that Face T can hide more sensitive info and mention that Face.evolve is a challenging model (and especially under such challenging models, the gap between w/ and w/o transformation is large). If needed, also discuss the poor result of face.evolve (w/o gt) + Face T in Table 2 and relate to the results here. All the discussions can go to appendix and be referred to in the main text.
% (btw, given that even with transformation, the final results of this case is still low, I'm wondering what will the results be if without the transformation. I kind of feel that transformation cannot very effectively solve the problem... not sure.)}

% \input{label_pred}

\noindent\textbf{Data Transformations.}\quad
Next, we discuss the performance of various data transformations on CelebA. 
We plot out label prediction accuracy w.r.t. $n$ for various transformations including the proposed sequential cutout, horizontal flipping, gray-scale, and color jittering in Fig.~\ref{fig:various_trans}.
We also plot the results without transformations.
We can see that sequential cutout performs better than other transformations in terms of label prediction accuracy.
Although it is also possible to adopt other transformations within our pipeline, it is non-trivial to select the best hyper-parameters for other transformations (\eg, cropping and color jittering).
We leave the analysis of how different transformations impact attack performance as future work.

\noindent\textbf{Overfitting Levels.}\quad
We also evaluate the impact of \textit{higher overfitting levels} of the target model on the performance of \method, since the overfitting phenomenon is key to model inversion attacks. 
Note that results reported in Table~\ref{tab:whitebox} and Table~\ref{tab:blackbox} are based on standard well-trained models.
We demonstrate that highly overfitted models are more vulnerable to our proposed attack.
We describe the relevant experiment setup and quantitative results in the appendix.

\vspace{-1em}
\section{Conclusion}
\vspace{-0.5em}
\looseness=-1
In this paper, we propose an effective private data recovery framework \method, which can effectively recover private information under a wide range of scenarios.
To our full knowledge, we are the first to propose an effective model inversion attack without prior knowledge of ground truth labels, which can achieve comparable results with previous methods that require ground truth labels.
If we are given such prior knowledge, we significantly outperform previous methods.
Our attack can also be applied under the $blackbox$ setting where the target model is provided as a service and not locally available.
We perform a comprehensive analysis of the performance of \method and our design choices.
We also demonstrate that our attack is robust against the purification defense.
We hope to raise people's concerns about possible negative effects of releasing pre-trained models online.
For future work, we are interested in whether we can perform privacy recovery simply with the target model and develop defenses against our attack.

\vspace{1em}
\noindent \textit{Acknowledgements.}
This work is partially supported by 
 NSF grant No.1910100, NSF CNS No.2046726, C3 AI, and the Alfred P. Sloan Foundation.

%%%%%%%%% REFERENCES
\clearpage
\bibliographystyle{splncs04}
\bibliography{egbib}

\clearpage
\appendix

\section{Additional Algorithm Details}

% \fan{I updated the pseudocode and the comments below. @Zhuowen, please do a careful check.}

% \fan{add a simple paragraph referring to the algorithm description in the main paper, and say that below we will concretely xxx}
In Section~\ref{sec:pseudo_label_predictor} and Section~\ref{sec:search}, we introduced \textbf{pseudo label predictor} and \textbf{latent vector selector} of \method. In this section, we concretely demonstrate the detailed algorithms for these two components.

\subsection{Detailed Algorithm for Pseudo Label Predictor}
\label{appx:pred_alg}
% \fan{Do not put the algorithm box alone. Add some text descriptions.}
Algorithm~\ref{alg:pred} elaborates how \method's pseudo label predictor infers a pseudo label for the target private image.
The pseudo label predictor randomly samples $n$ latent vectors from the prior distribution of the generation backbone iteratively. For each latent vector, we perform $m$ transformations to the recovered image.
We compute the average confidence for each class label $c \in [1, C]$ on the recovered image and their transformed versions.
Then we select the class label $\hat c$ with the highest average confidence as the predicted pseudo label.

\begin{algorithm}
\caption{Label Prediction}
\label{alg:pred}
\begin{algorithmic}

\State \textbf{Input:} $x_{ns}$: corrupted private data, $\mathcal {T} = \{t_i\}_{i=1}^{m}$: $m$ transformations, $F$: target model, $n$: number of randomly sampled latent vectors, $C$: number of classes 
% \fan{remember to change to the updated notations}

\State $\bm{v} = \bm{0}$ \Comment{\textit{\small Initialize $\bm{v}$ to zero vector of length $C$}}
\For{$i \gets 1$ to $n$}
    \State $z_i \sim \mathcal{N}(0, 1)$ \Comment{\textit{\small Randomly sample one latent vector from Gaussian}}
    \State $\tilde x_i^0 \gets G(z_i, x_{ns})$ \Comment{\textit{\small Generate the recovered image with the generation backbone}}
    \State $\tilde x_i^j \gets t_j(\tilde x_i^0)$ for $j=1\ldots m$ \Comment{\textit{\small Generate $m$ transformed images}}
    \State $ \bm{v} \gets\bm{v} +\sum_{j=0}^{m} F(\tilde x_i^j)$ \Comment{\textit{\small Sum over the prediction vectors of the $m$ transformations}}
\EndFor
\State $\bm{v} \gets  \frac{1}{ n(m+1)} \bm{v}$
\State $\hat c \gets \arg\max_{c\in[1,c]}\bm{v}_c$ \\\Comment{\textit{\small Select the class $c$ that has the highest value under measure $\cal M$ (Eq.~\ref{eq:metric})}}

\State \textbf{return} $\hat c$

\end{algorithmic}
\end{algorithm}

% $\mathcal {T} = \{t_i\}_{i=1}^{m}$.
% On each recovered image $\hat x_i$, we  perform $m$ transformations independently to obtain $m$ transformed images $\{\tilde x_i^j\}_{j=1}^{m}$, where $\tilde x_i^j=t_j(\hat x_i)$.
% We additionally define $\tilde x_i^0=\hat x_i$.

\subsection{Detailed Algorithm for Latent Vector Selector}
\label{appx:latent-sel-algo}
Algorithm~\ref{alg:select} describes how we select the latent vector given the target label.
We refer to the target label as the predicted pseudo label or ground truth label (if available).
We first randomly sample $n$ latent vectors and filter out the latent vectors that can generate recovered images that will be predicted as the target label by the target model.
If there exist such latent vectors, we choose the latent vector which leads to the recovery with the highest confidence of the target label; otherwise, the algorithm returns a random latent vector sampled from Gaussian.

\begin{algorithm}
\caption{Latent Vector Selection}
% \fan{similar comments as for Algorithm 1} \bl{maybe leave this one here}
\label{alg:select}
\begin{algorithmic}
\small
\State \textbf{Input:} $x_{ns}$: corrupted private data, $F$: target model, $n$: number of randomly sampled latent vectors, $y$: predicted label or ground truth label

\State $z_{bank} \gets $ sample $n$ latent vectors from $\mathcal{N}(0, 1)$
\State $z_{bank\_y} \gets $ $\{z \in z_{bank} | \arg\max_{c\in[1,C]} F_c(G(z, x_{ns}))=y\} $ 
\State \Comment{\textit{\small Select latent vectors that can generate recovered images that will be predicted to have label $y$}}
% iterate over $z_{bank}$, select $z \in z_{bank}$ where $\arg\max_{c\in[1,C]} F_c(G(z, x_{ns}))=y$
\If{$z_{bank\_y}$ is not empty}
    % \State \sout{$c_{y} \gets F_y(G(z_{bank\_y}, x_{ns}))$ for $z \in z_{bank\_y}$}
    % \State \sout{$\mathit{index} \gets \arg\max c_{y}$}
    % \State \sout{$\hat z \gets c_{y}[\mathit{index}]$}
    \State $\hat z \gets \arg\max_{z \in z_{bank\_y}} F_y(G(z, x_{ns}))$ \Comment{\textit{\small Select the latent vector that has the highest confidence for label $y$}}
\Else
    % \State \sout{$z \sim \mathcal{N}(0, 1)$}
    % \State \sout{$\hat z \gets z$}
    \State $\hat z \sim \mathcal{N}(0, 1)$
\EndIf

\State \textbf{return} $\hat z$

\end{algorithmic}
\end{algorithm}

\subsection{\method with Known Ground Truth Labels}
\label{appx:known-gt}
\begin{figure}[t]
    \centering
    \includegraphics[width=.8\textwidth]{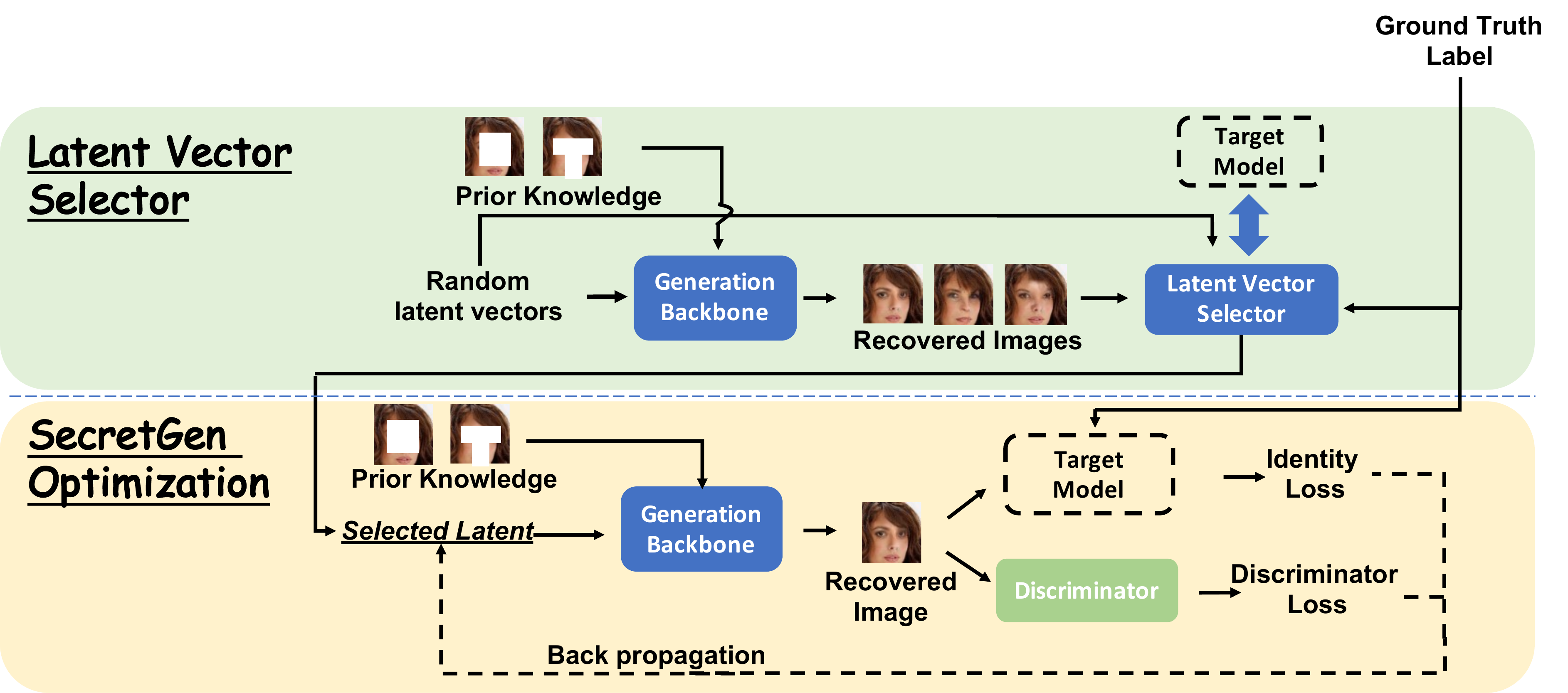}
    \caption{\small Overview of the proposed \method if ground truth label is available. The \textcolor{algocolor}{\textbf{blue modules}} represent the proposed algorithms. The \textit{Target Model} could allow either whitebox or blackbox access. The ground truth label is fed to the latent vector selector, which returns the initial latent vector for \method optimization. In the whitebox setting, the ground truth label is also leveraged for identity loss.}
    \label{fig:gt_case}
\end{figure}

\looseness=-1
As we have presented in Section~\ref{sec:method}, \method predicts a pseudo label for the target private image through a discrimination metric without requiring the knowledge of the ground truth label. Here, we consider the setting when the ground truth label is available, and we point out that the knowledge can be incorporated into our pipeline conveniently.
See Fig.~\ref{fig:gt_case} for an overview of \method if ground truth labels for target private images are known.
Under this scenario, the pseudo label predictor is not required, and we substitute the input of the latent vector selector with the known ground truth label. 
% \chaowei{Give more details. How it works.}
Besides, the ground truth label is leveraged for computing identity loss during \method optimization in the whitebox setting.

\section{Additional Experimental Setup}
\label{appx:details}
\noindent\textbf{Target Model Training.}\quad
The target models are trained on $D_{\rm pri}^{\rm train}$.
For training \texttt{ViT-B\_16}, we use the SGD optimizer with learning rate $3 \times 10^{-2}$, batch size 64 for 10,000 steps.
We use cosine annealing schedule with 500 warm-up steps.
The other models are trained with the SGD optimizer with learning rate $10^{-2}$, batch size 64, momentum 0.9 and weight decay $10^{-4}$.
% The classification accuracy on private identities of the resulting target models are: 74.72\% (\texttt{VGG16}), 67.46\% (\texttt{ResNet152}), 96.9\% (\texttt{face.evoLVe}) and 63.9\% (\texttt{ViT-B\_16}). 

\noindent\textbf{Attack Procedures.}\quad
We leverage WGAN-GP~\cite{gulrajani2017improved} and train our \textit{generation backbone} with the Adam optimizer~\cite{kingma2014adam} with batch size 64, learning rate $4\times 10^{-3}$, $\beta_1=0.5$, $\beta_2=0.999$.
We set $\lambda_{\rm div}$ in Eqn.~\eqref{eq:gen-backbone} to 0.5 based on quantitative observations.
For both our \textit{pseudo label predictor} and \textit{latent vector selector}, we set the number of randomly sampled latent vectors to 200.
Although we demonstrate that our pipeline works better under larger $n$, the efficiency suffers.
See Section~\ref{sec:pred_acc} for concrete results.
We set the patch size for cutouts to $16 \times 16$ and the resulting number of transformations $m$ is 16.
See Appendix~\ref{appx:m} for more discussion regarding the selection of $m$.
At the \textit{\method optimization} stage, we set $\lambda_{\rm id}$ in Eqn.~\eqref{eq:opt-main} to 100 and use the SGD optimizer to optimize the selected latent vector for 1,500 iterations with learning rate 0.02 and momentum 0.9.

\section{Additional Results}

% \chaowei{The titles of Section B and C are a little confused. Why we need to have a new section? We could gather the description of overfitting in Section B and C . }
In this section, we report additional results for ablation studies. We also show some additional qualitative results of \method.
% in this section are conducted on a portion of private training data, which consists of 3,200 randomly sampled images from $D_{\rm pri}^{\rm train}$ due to limited time and resources.

\subsection{Attack Performance on FaceScrub}
Table~\ref{tab:whitebox-fs} compares attack performance of \method and baselines on FaceScrub. We use \texttt{VGG16} as the target model.
It can be seen that \method still significantly outperforms PII and GMI.
The conclusion aligns with that on CelebA.

\begin{table}[ht]
    \centering
    \caption{\small Whitebox attack performance of \method and baselines on FaceScrub. Target model: \texttt{VGG16}. The given prior information is corrupted images by center mask.}
    
    \label{tab:whitebox-fs}
    \resizebox{\textwidth}{!}{
    \begin{tabularx}{\linewidth}{cc>{\centering\arraybackslash}X>{\centering\arraybackslash}X>{\centering\arraybackslash}X}
    \toprule
\textbf{Method} & \textbf{~Ground Truth Label~} & \textbf{Protocol 1} & \textbf{Protocol 2} & \textbf{PSNR} \\ \hline
PII & \xmark & 0.274 & 0.818 & 26.269 \\
GMI & \cmark & 0.398 & 0.972 & 26.028 \\
SecretGen & \xmark & 0.396 & 0.942 & 26.459 \\
SecretGen & \cmark & \textbf{0.453} & \textbf{0.973} & \textbf{26.516} \\
\bottomrule
\end{tabularx}
    }
\end{table}

\subsection{Ablation Studies on the Number of Transformations}
\label{appx:m}
\begin{figure}[ht]

\newlength{\utilheightfull}
\settoheight{\utilheightfull}{\includegraphics[width=.3\linewidth]{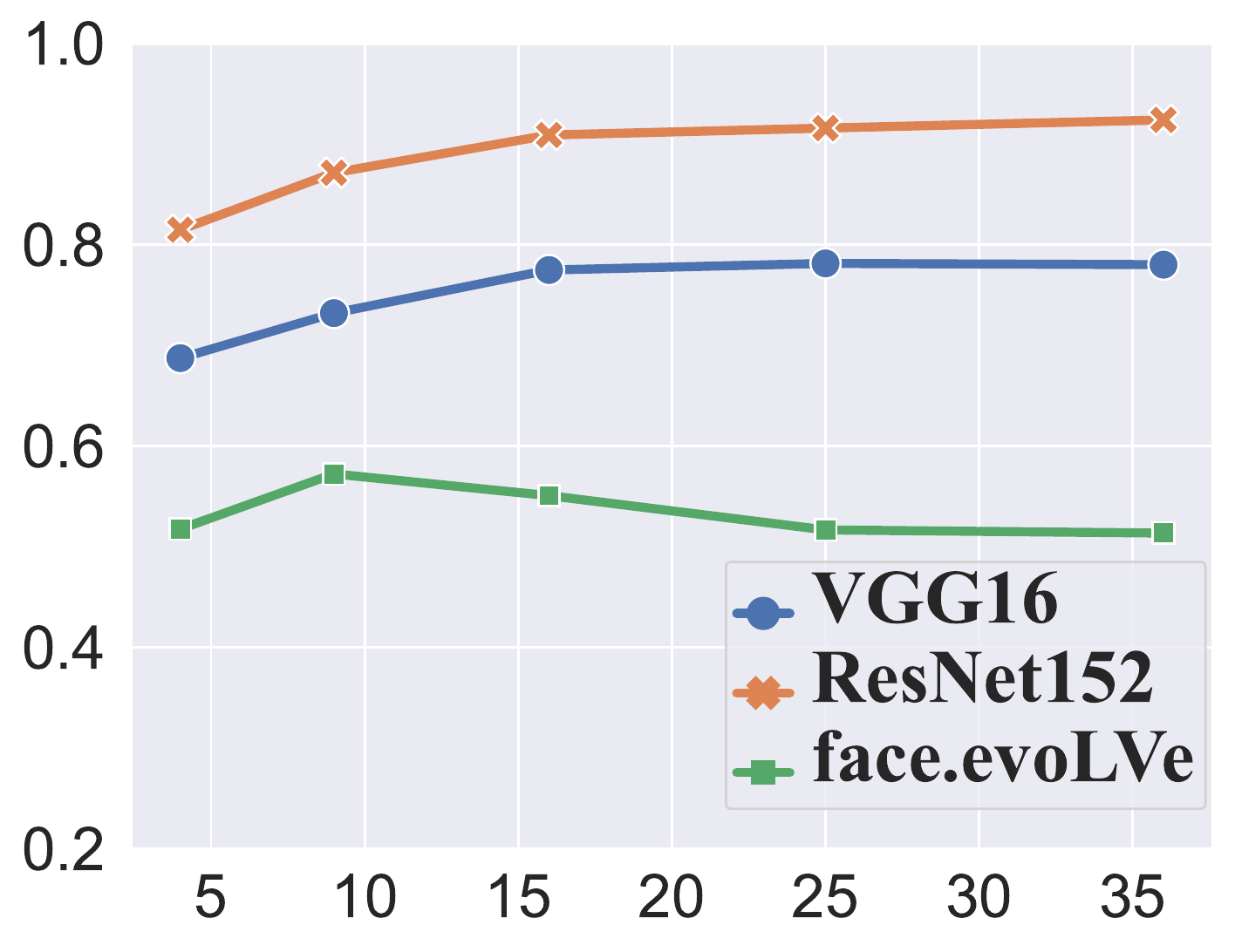}}%

\newcommand{\rowname}[1]% #1 = text
{\rotatebox{90}{\makebox[\utilheightfull][c]{\tiny #1}}}

\centering

{
\renewcommand{\tabcolsep}{10pt}

\begin{subtable}
\centering
\resizebox{0.6\linewidth}{!}{%
\begin{tabular}{@{}p{6mm}@{}c@{~~~~~~~~}c@{}}
        % & \makecell{\small{\textbf{VGG16}}}
        & \makecell{~~~~\small{\textbf{Center Mask}}}
        & \makecell{~~~~\small{\textbf{Face T Mask}}}\\
\rowname{\small Accuracy }&
\includegraphics[height=\utilheightfull]{figs/m_center.pdf}& 
\includegraphics[height=\utilheightfull]{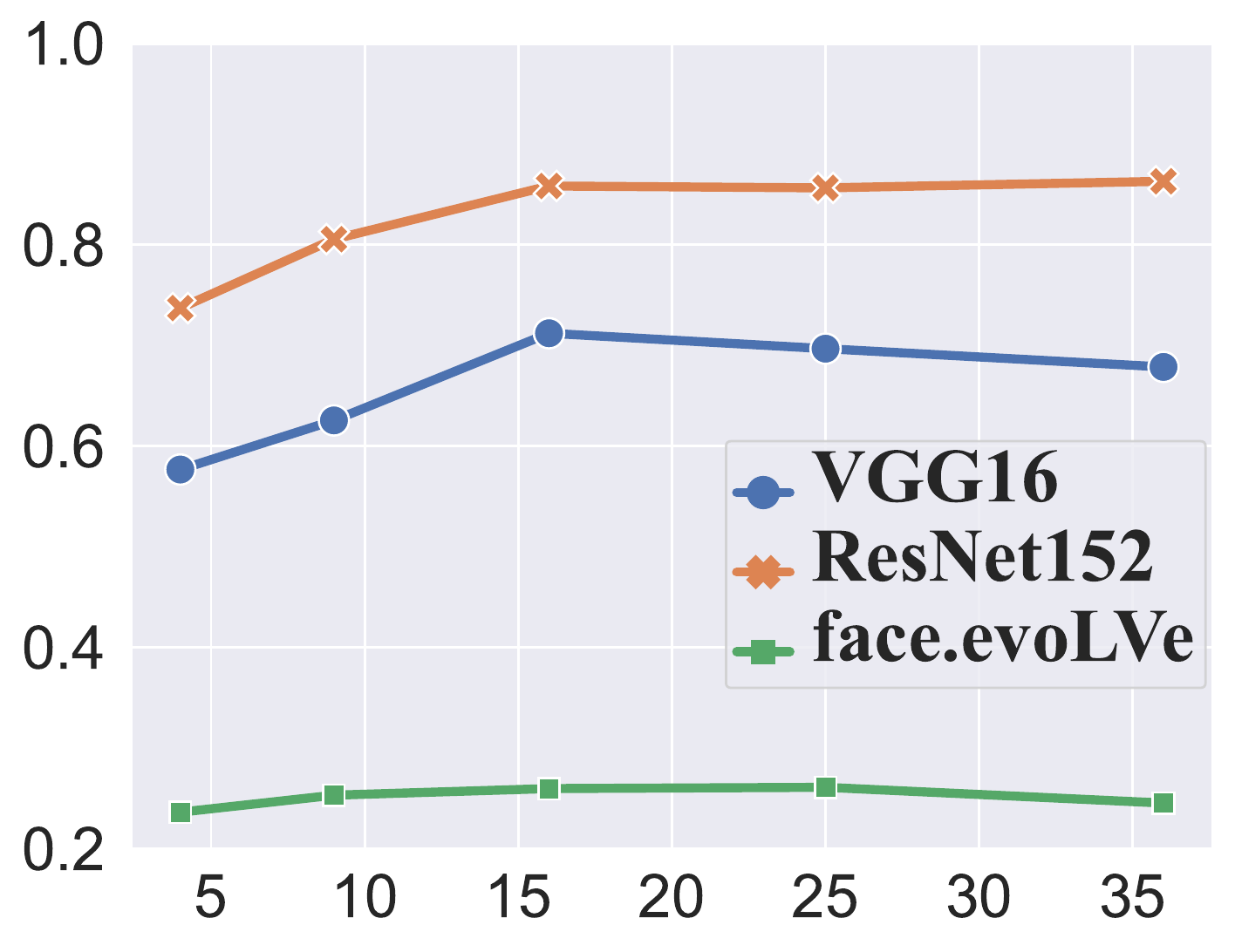}
\\[-1ex]
        & \makecell{~~~~\small{$m$}}
        & \makecell{~~~~\small{$m$}}
\end{tabular}
}
% \caption{\small Certified radius $R_t$ along time steps}\label{tab:cert-rad}
\end{subtable}

}
\vspace{-3mm}
\caption{\small Label prediction accuracy of the pseudo label predictor w.r.t. different number of transformations on CelebA. For each private image, we sample $n=200$ latent vectors, and then perform $m$ transformations corresponding to each latent vector. }
\label{fig:m_full}
\vspace{-2em}
\end{figure}

% \looseness=-1
We demonstrate that the number of transformations $m$ in the discrimination metric (Eqn.\eqref{eq:metric}) affects the performance of our pseudo label predictor.
We adjust $m$ by changing the patch size of our sequential cutout.
Since our recovered image is $64 \times 64$, the resulting patch size is thus $\frac{64}{\sqrt{m}} \times \frac{64}{\sqrt{m}}$.
We fix $n$ to 200 and plot label prediction accuracy on the recovered images w.r.t. $m$ in Fig.~\ref{fig:m_full}.

% \looseness=-1
We observe that when $m$ is small, the resulting label prediction accuracy suffers due to the aggressiveness of cutting out a large patch.
However, efficiency suffers as $m$ increases.
See Table~\ref{tab:eff} for the average running time of pseudo label predictor for one target private image on a single NVIDIA RTX 2080 Ti GPU.
Note that the relationship between $m$ and label prediction accuracy is not consistent along different model architectures.
Since the attacker has no idea about the optimal $m$ for the target model, in practice we set $m$ to 16 and we demonstrate that performance of \method is promising in Section~\ref{sec:exp}.

% Therefore, it's non-trivial to select the optimal $m$ since the attacker has no idea which $m$ works the best during attack.
% We set $m$ to $16 \times 16$ in our pipeline as a general choice for all model architectures to balance performance and efficiency.

\renewcommand{\thesubfigure}{\alph{subfigure}}

\begin{table}[h]
    \centering
    \caption{\small Average running time of pseudo label predictor for one target private image. For each private image, we sample $n=200$ latent vectors and then perform $m$ transformations corresponding to each latent vector. Target model: \texttt{VGG16}. }
    
    \resizebox{\textwidth}{!}{
    \begin{tabularx}{\linewidth}{c>{\centering\arraybackslash}X>{\centering\arraybackslash}X>{\centering\arraybackslash}X>{\centering\arraybackslash}X>{\centering\arraybackslash}X}
    \toprule
    \textbf{m} & 4 & 9 & 16 & 25 & 36 \\ 
    \midrule
    \textbf{Running Time (s)} & ~18.583~ & ~26.747~ & ~38.222~ & ~53.463~ & ~72.876~ \\
    \bottomrule
    \end{tabularx}
    }
    \label{tab:eff}
\end{table}

\subsection{Ablation Studies on Transformations}
\label{appx:ablation}
% In this subsection, we conduct ablation studies for the elements involved in our component Pseudo Label Predictor, including \textit{the number of transformations} and \textit{the transformation types}.
In this section, we present a series of quantitative results of the ablation studies on transformations.

% We compare $\mathcal{M}$ with $\mathcal{M}'$ on \textbf{attack accuracy} under Protocol 1 and \textbf{label prediction accuracy} which quantifies the performance of our pseudo label predictor.
% \fan{Basically you introduce two experiments of different purposes here, but this is not very clear. Make the \textbf{experiment purpose}, \textbf{experiment setting}, and \textbf{conclusions} more clear. Maybe can separate into two paragraphs?}

\noindent\textbf{Performance without Transformations.}\quad
In Section~\ref{sec:pred_acc}, we demonstrated that \method performs better if transformations are incorporated in the discrimination metric for \texttt{face.evoLVe}. 
Here we present the results for other model architectures.
We remove transformations from the original discrimination metric $\mathcal{M}$ (Eqn.~\eqref{eq:metric}) and denote the resulting discrimination metric without transformations as $\mathcal{M}'$ (Eqn.~\eqref{eq:m2}).
We evaluate \textit{end-to-end} attack accuracy for $\mathcal{M}$ and $\mathcal{M}'$ on the first 3,200 images from $D_{\rm pri}^{\rm train}$ under Protocol 1. 
The results are illustrated in Table~\ref{tab:trans_full}, which demonstrate that incorporating transformations improves attack accuracy for \texttt{VGG16}.
For \texttt{ResNet152}, the results are similar.
We also report PSNR scores for the recovered images for reference.

\begin{figure}[ht]

\newlength{\utilheightpred}
\settoheight{\utilheightpred}{\includegraphics[width=.3\linewidth]{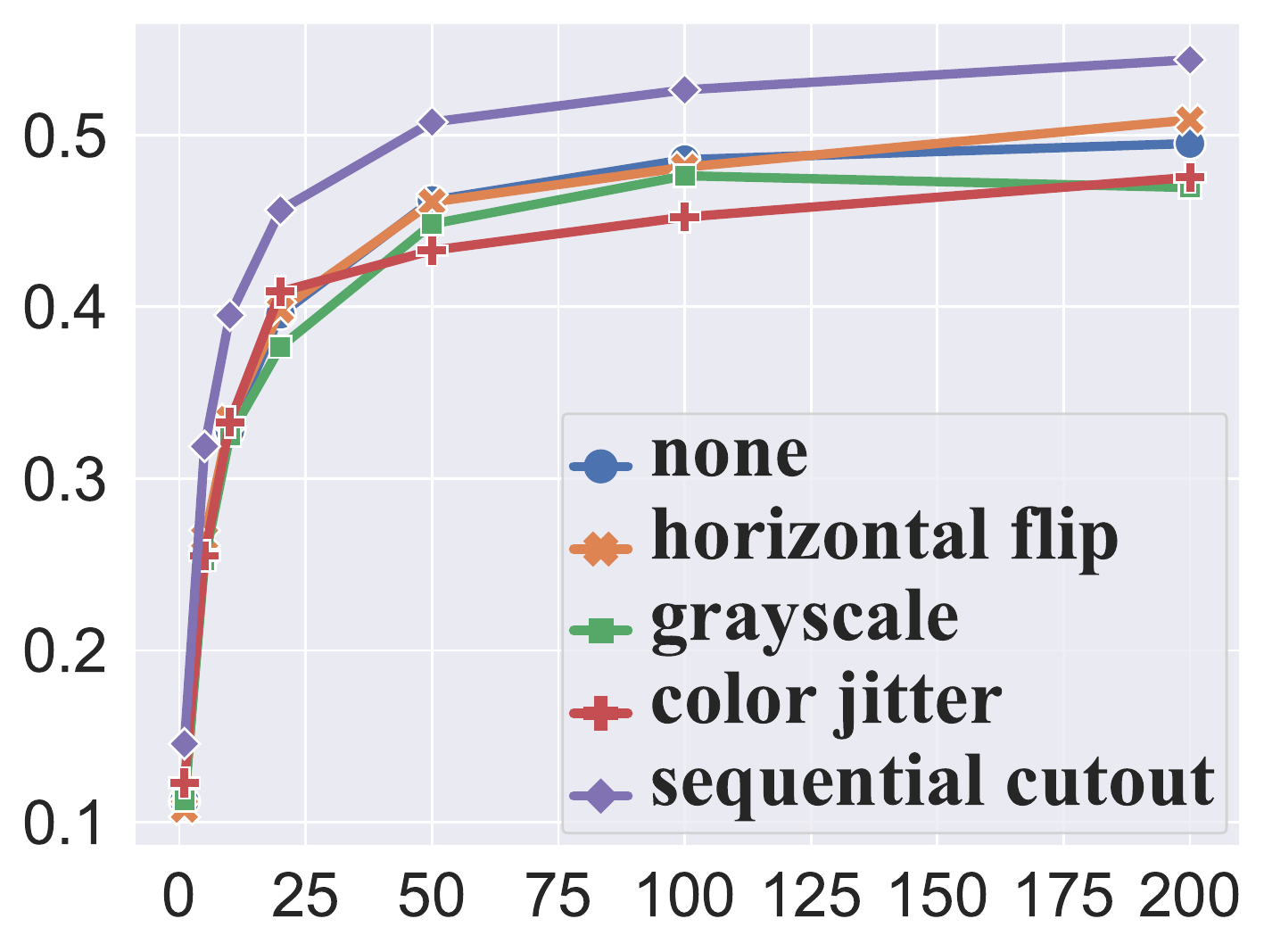}}%

\newcommand{\rowname}[1]% #1 = text
{\rotatebox{90}{\makebox[\utilheightpred][c]{\tiny #1}}}

\centering

{
\renewcommand{\tabcolsep}{10pt}

\begin{subtable}
\centering
\resizebox{0.85\linewidth}{!}{%
\begin{tabular}{@{}p{8mm}@{}c@{}c@{}c@{}}
        & \makecell{\small{\textbf{~~~~VGG16}}}
        & \makecell{\small{\textbf{~~~~ResNet152}}}
        & \makecell{\small{\textbf{~~~~face.evoLVe}}}\\
        % & \makecell{\Large{\textbf{VGG16}}}
\rowname{\makecell{\small \textbf{Center Mask} \\\small Accuracy }}&
% \rowname{\footnotesize Accuracy }&
\includegraphics[height=\utilheightpred]{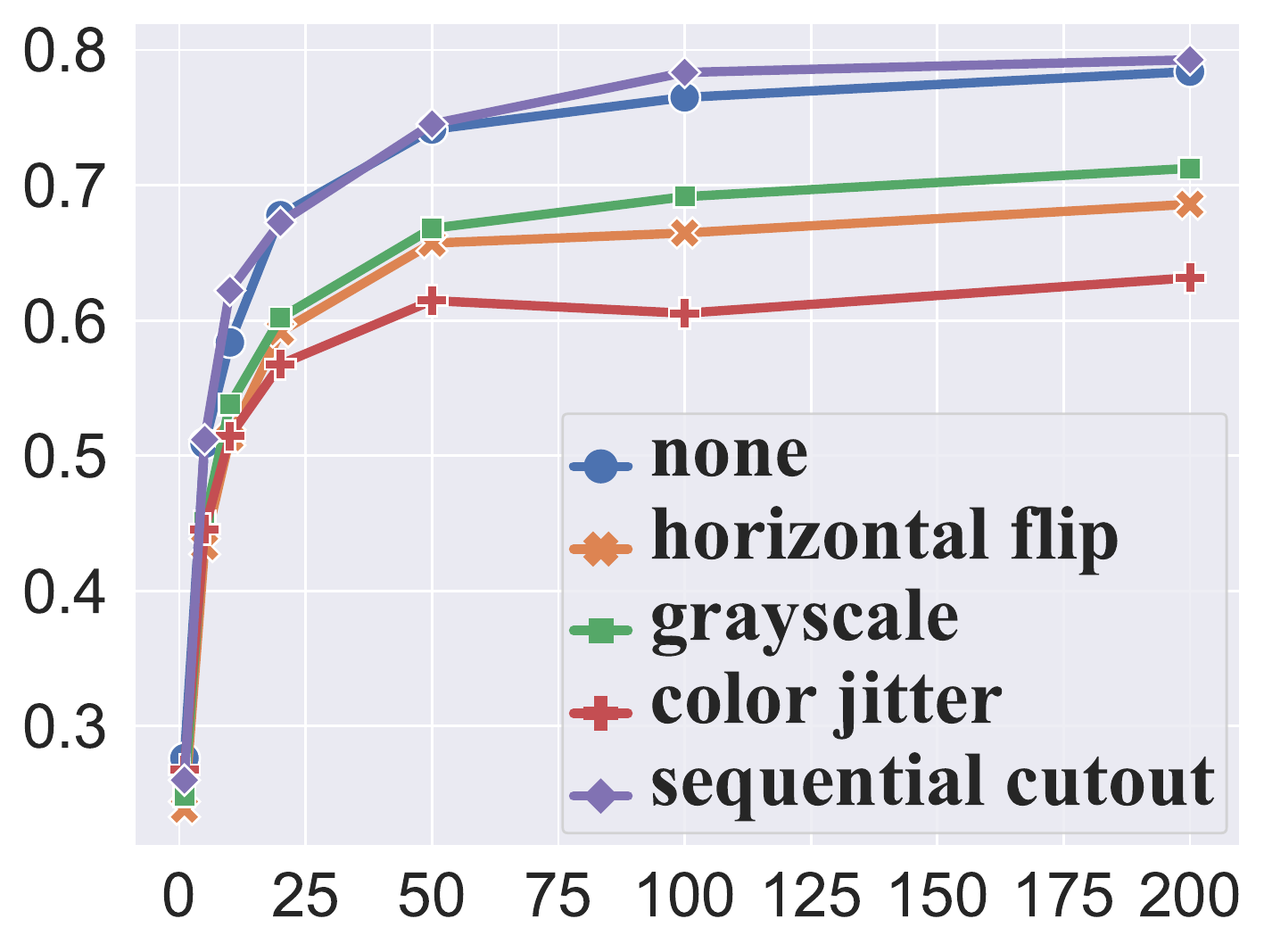}& 
\includegraphics[height=\utilheightpred]{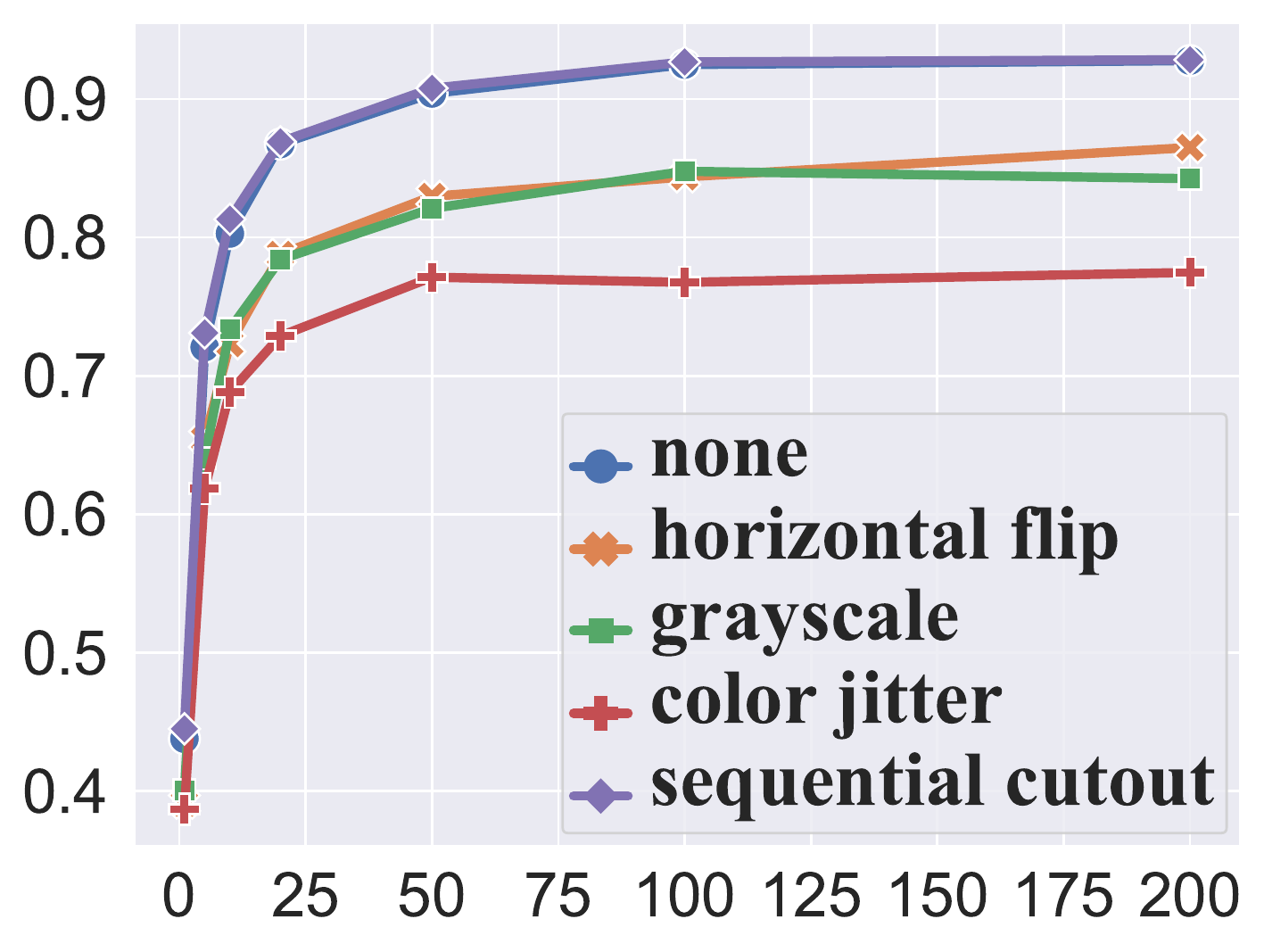}& 
\includegraphics[height=\utilheightpred]{figs/variance_ir50_center.pdf}\\
\rowname{\makecell{\small \textbf{Face T Mask} \\\small Accuracy }}&
\includegraphics[height=\utilheightpred]{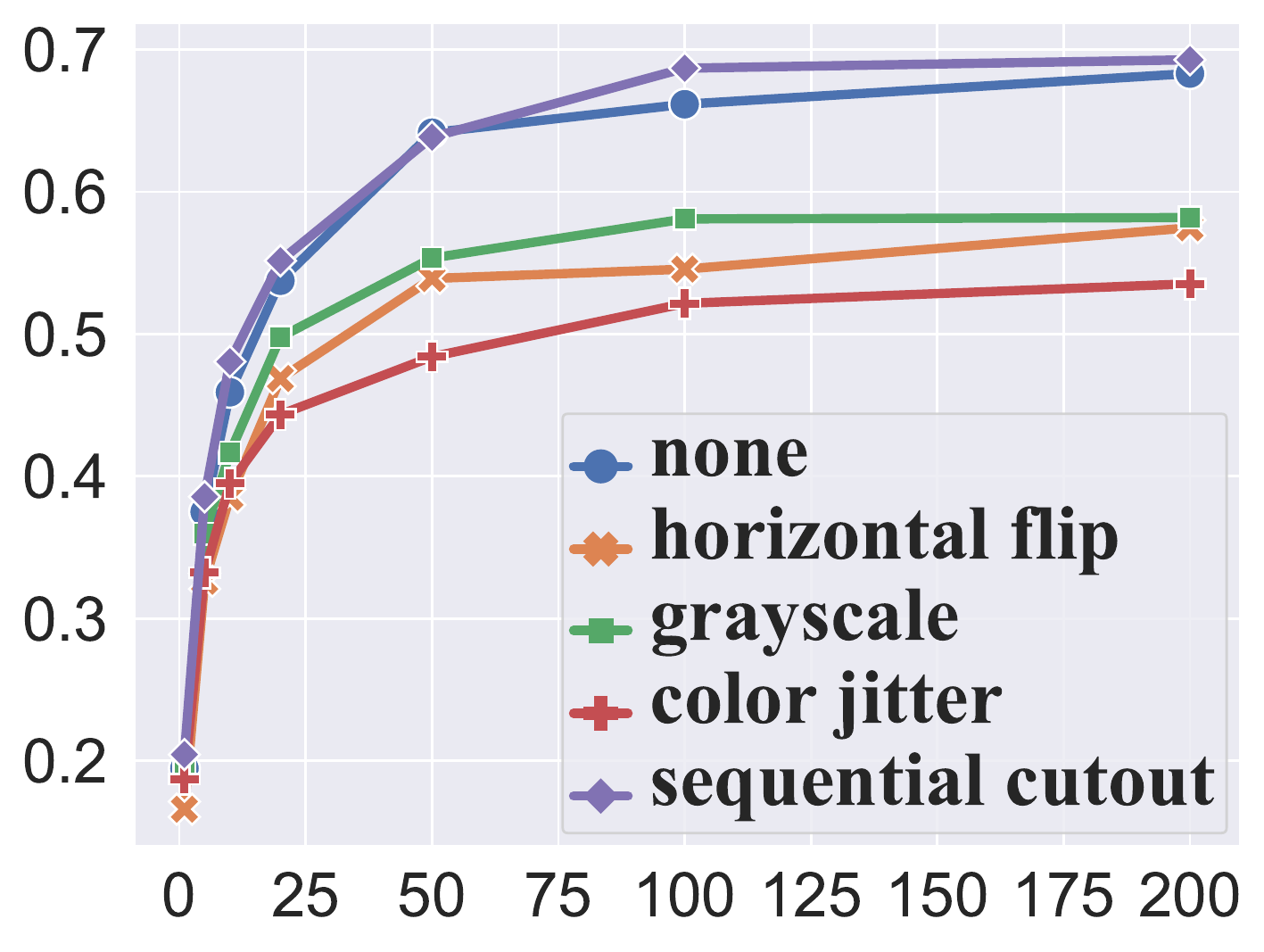}&
\includegraphics[height=\utilheightpred]{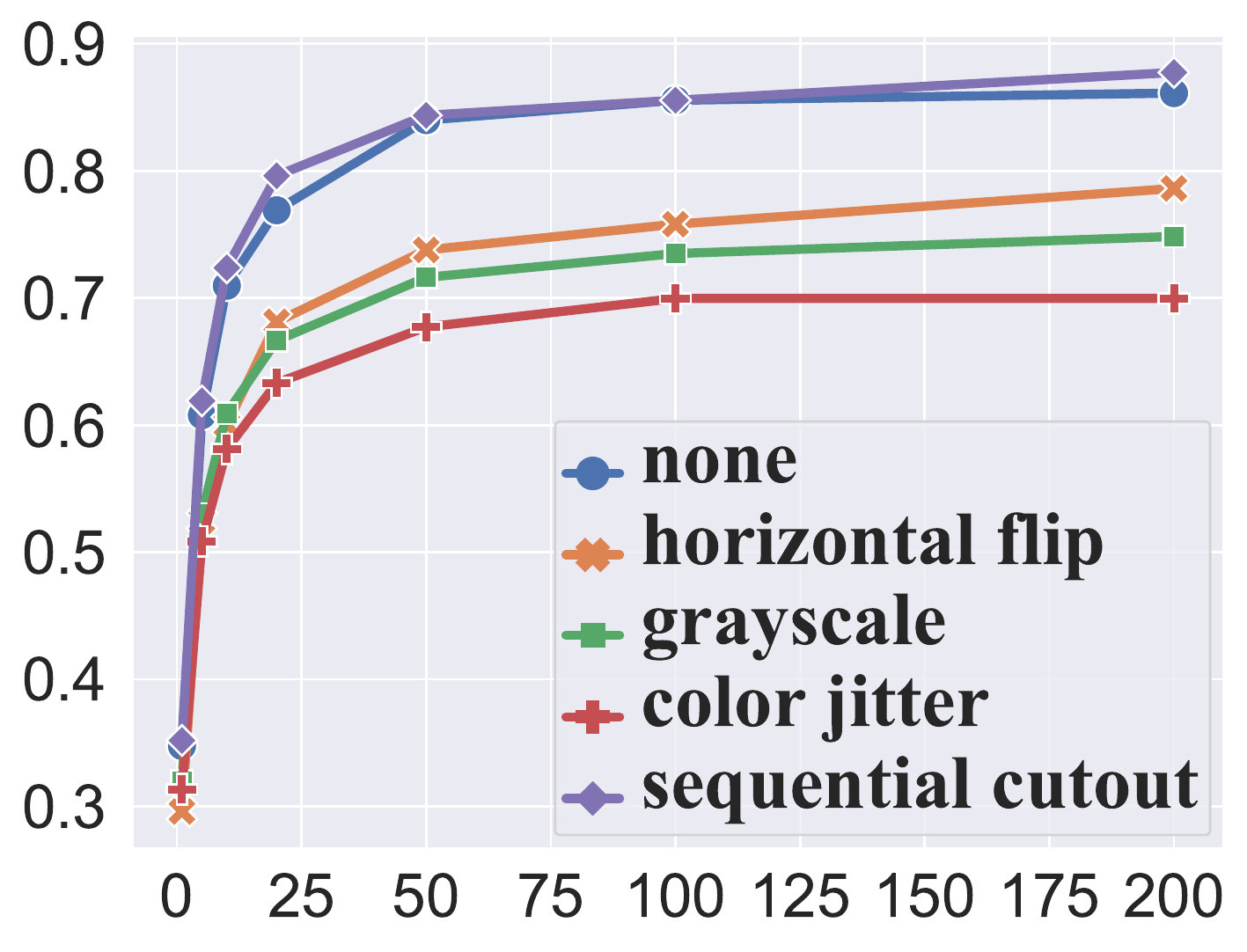}&
\includegraphics[height=\utilheightpred]{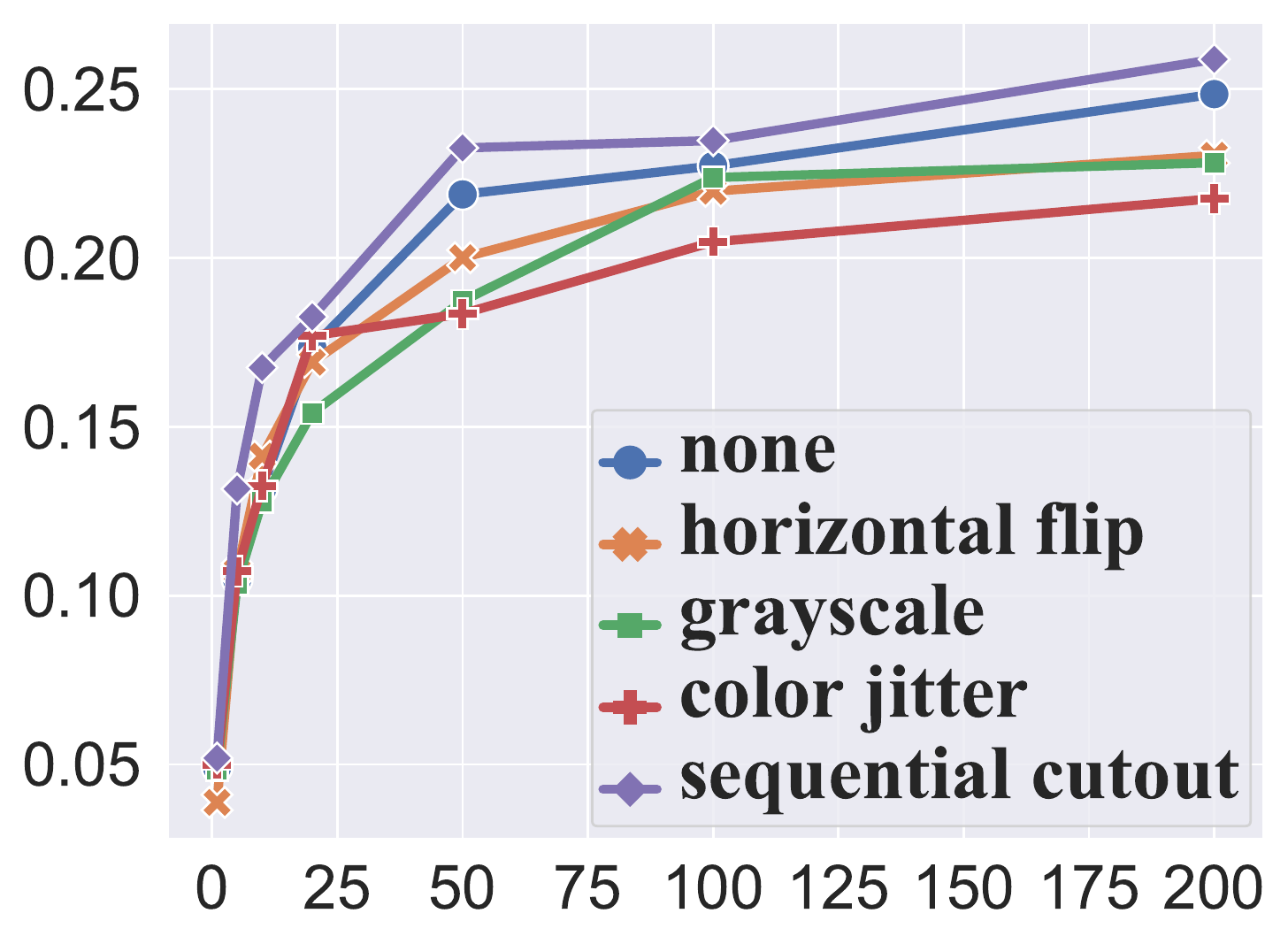}\\
\\[-3ex]
        & \makecell{~~~~\large{$n$}}
        & \makecell{~~~~\large{$n$}}
        & \makecell{~~~~\large{$n$}}
\end{tabular}
}
% \caption{\Large Certified radius $R_t$ along time steps}\label{tab:cert-rad}
\end{subtable}

}

\caption{\small Label prediction accuracy without transformation (denoted as ``none'' in legend) and with different kinds of transformations on CelebA. We plot label prediction accuracy w.r.t. the number of randomly sampled latent vectors.}
\label{fig:trans_full}

\end{figure}

\noindent\textbf{Performance with Various Transformations.}\quad
We further evaluate the performance of our pseudo label predictor incorporated with different kinds of transformations.
% We evaluate label prediction accuracy to demonstrate the impact of transformations on pseudo label predictor.
% Intuitively, if the pseudo label predictor infers labels more correctly, the \textit{end-to-end} attack accuracy improves.
We consider using more aggressive transformations including horizontal flipping, grayscale, and color jittering (brightness 1.5, contrast 1.5, saturation 1.5).
We plot label prediction accuracy for the first 3,200 private images from $D_{\rm pri}^{\rm train}$ with different transformations w.r.t. the number of random latent vectors in Fig.~\ref{fig:trans_full}.
We also plot the label prediction accuracy without transformations for reference. 
We demonstrate that the performance of aggressive transformations is worse than the case when no transformations are incorporated, while sequential cutout which is used in our pipeline performs the best.
% Intuitively, sequential cutout keeps most of $x_{ns}$ unchanged, while in other transformations $x_{ns}$ is also affected.

\begin{table}
\centering
\caption{\small Attack accuracy of \method with and without transformations on CelebA. Evaluated with the evaluation model under Protocol 1.}

\resizebox{\linewidth}{!}{
\begin{tabular}{cccccc}
\toprule
\multirow{2}{*}{\textbf{Target Model}} & \multirow{2}{*}{\textbf{Setting}} & \multicolumn{2}{c}{\textbf{Center Mask}} & \multicolumn{2}{c}{\textbf{Face T Mask}} \\ \cline{3-6} 
                                       &                                   & \textbf{Attack Acc}   & \textbf{PSNR}    & \textbf{Attack Acc}    & \textbf{PSNR}   \\ \midrule
\multirow{2}{*}{VGG16}                 & w/o transformation                & 0.584                 & 27.863           & 0.319                  & 26.596          \\
                                       & w/ transformation                 & \textbf{0.599}        & \textbf{27.874}  & \textbf{0.328}         & \textbf{26.598}          \\ \hline
\multirow{2}{*}{ResNet152}             & w/o transformation                & \textbf{0.594}                 & \textbf{27.447}           & 0.337                  & \textbf{26.792}          \\
                                       & w/ transformation                 & 0.592                 & 27.442           & \textbf{0.338}                  & 26.788          \\ \hline
\multirow{2}{*}{face.evoLVe}           & w/o transformation                & 0.528                 & 27.505           & 0.256                  & 26.527          \\
                                       & w/ transformation                 & \textbf{0.550}        & \textbf{27.522}  & \textbf{0.273}         & 26.527  \\ 
\bottomrule
\end{tabular}
}
\label{tab:trans_full}
\end{table}

\renewcommand{\thesubfigure}{\alph{subfigure}}

\subsection{Impact of Overfitting Levels}
\label{sec:overfit_results}
\looseness=-1
We evaluate target models of two architectures: \texttt{VGG16} and \texttt{ResNet152}.
We train the target model from scratch on $D_{\rm pri}^{\rm train}$ under the settings in Section~\ref{sec:setup} for 200 epochs and we save the model at epoch 50, 100, 200.
Classification accuracy of overfitted models of the same architecture are similar, which are 0.746, 0.748, 0.748 (\texttt{VGG16}) and 0.675, 0.676, 0.675 (\texttt{ResNet152}), respectively. 
We substitute the target model with a public feature extractor from \cite{cheng2017know} when training the generation backbone to avoid training multiple generation backbones for each model.

\begin{figure}[ht]

\newlength{\utilheightoverfit}
\settoheight{\utilheightoverfit}{\includegraphics[width=.3\linewidth]{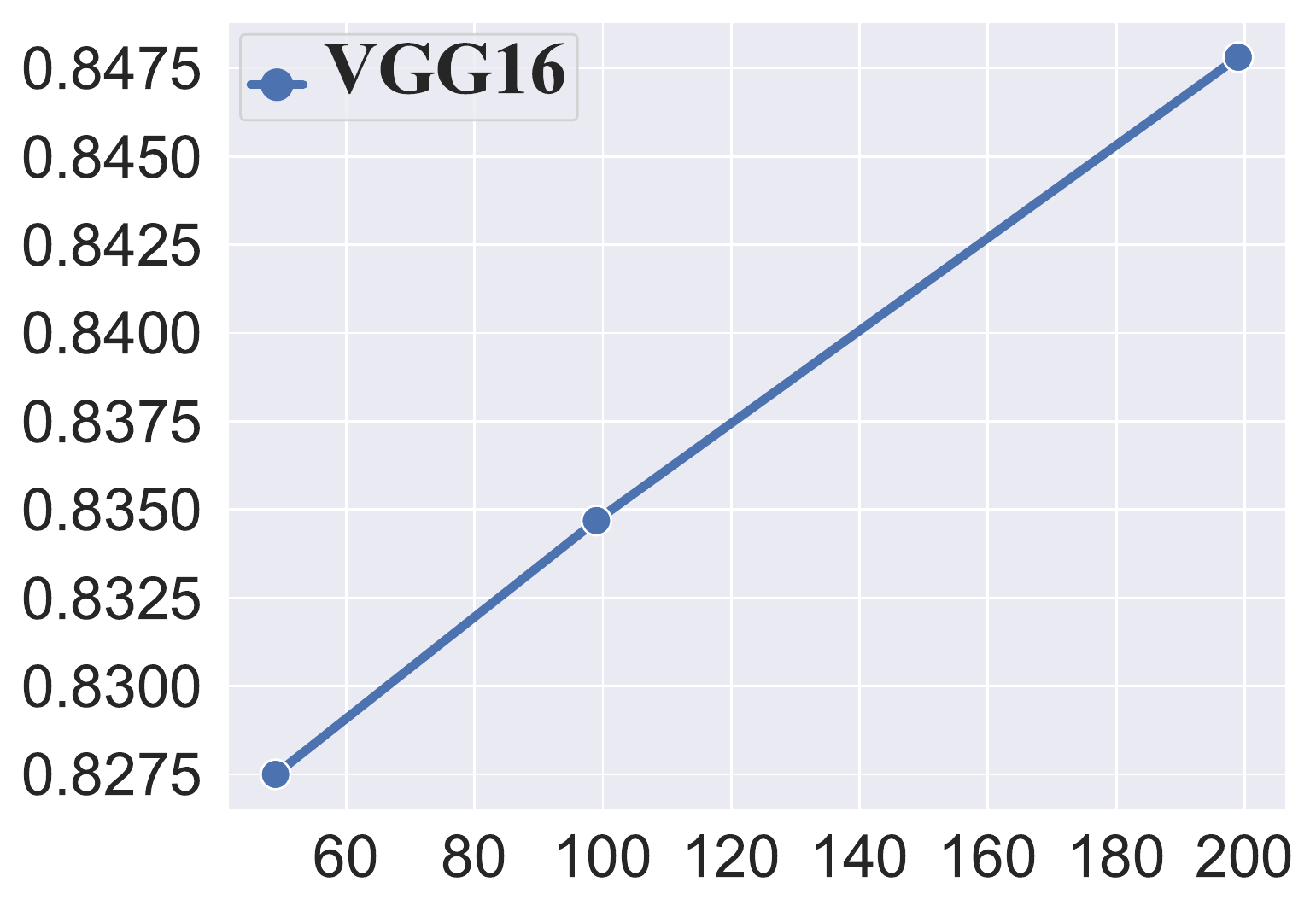}}%

\newcommand{\rowname}[1]% #1 = text
{\rotatebox{90}{\makebox[\utilheightoverfit][c]{\tiny #1}}}

\centering

{
\renewcommand{\tabcolsep}{10pt}

\begin{subtable}
\centering
\resizebox{0.6\linewidth}{!}{%
\begin{tabular}{@{}p{6mm}@{}c@{~~~~~~~~}c@{}}
        % & \makecell{\small{\textbf{VGG16}}}
        & \makecell{\small{~~~~\textbf{Center Mask}}}
        & \makecell{\small{~~~~\textbf{Face T Mask}}}\\
\rowname{\small Accuracy }&
\includegraphics[height=\utilheightoverfit]{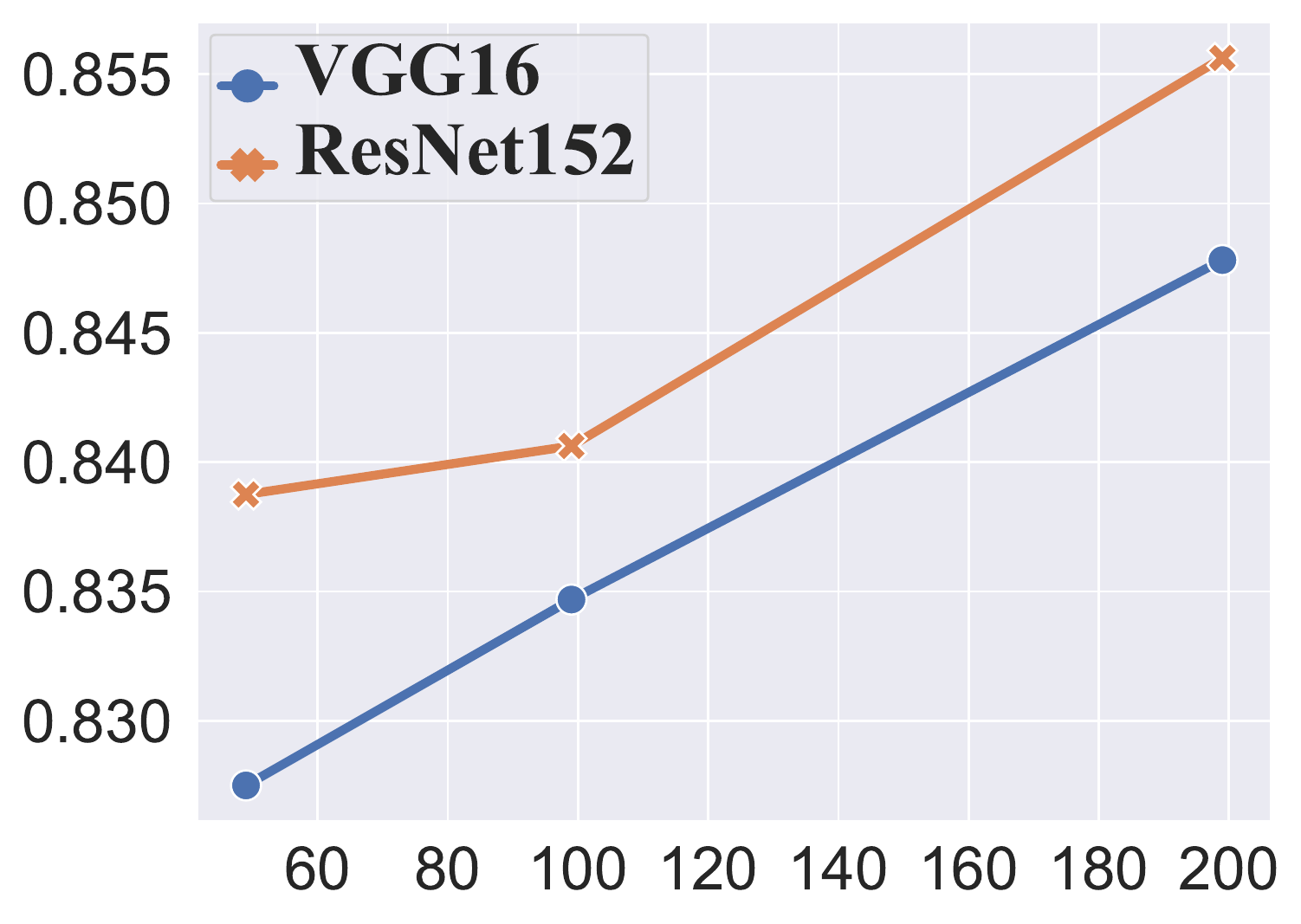}& 
\includegraphics[height=\utilheightoverfit]{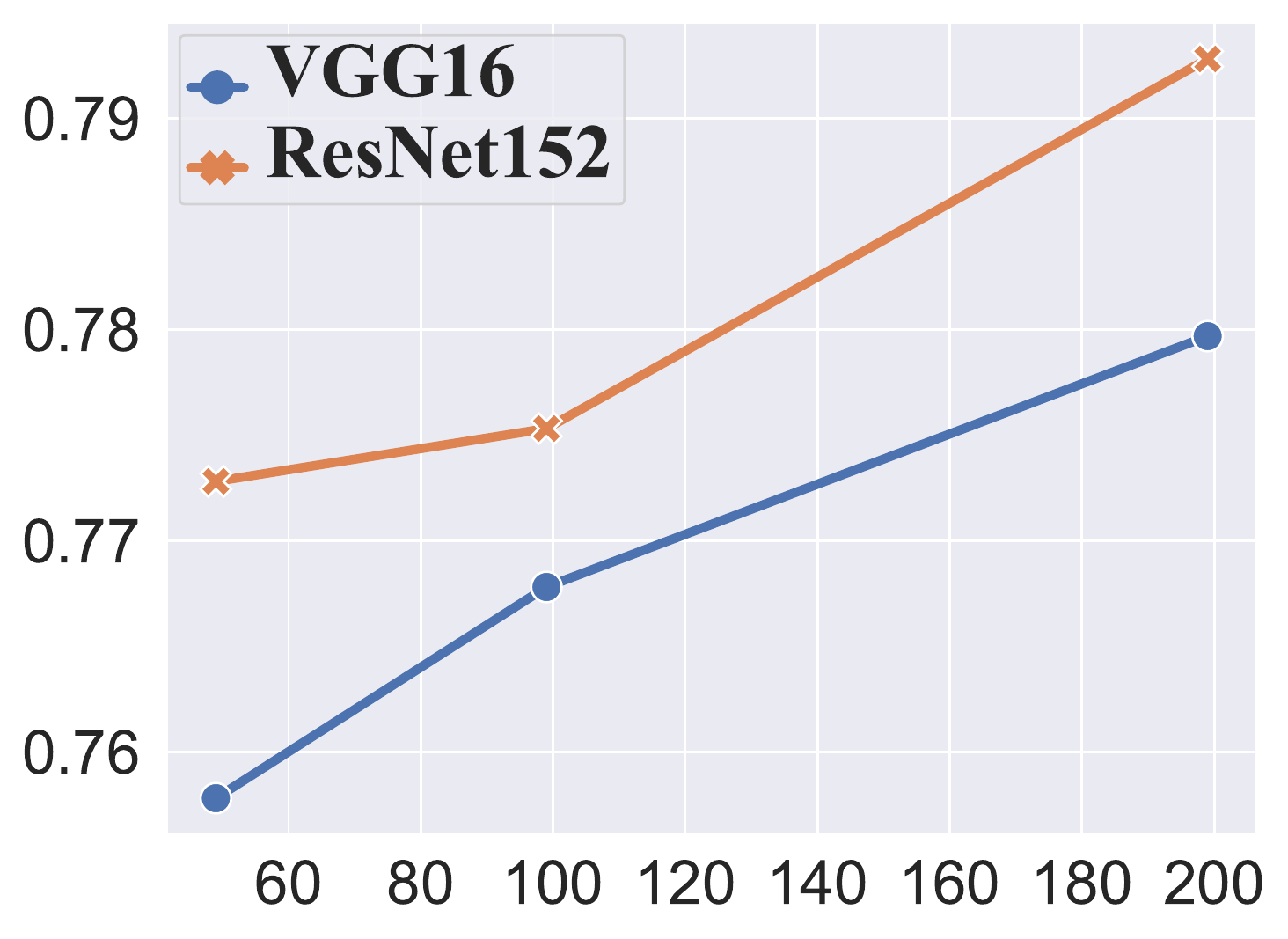}
\\[-1ex]
        & \makecell{~~~~\small{epoch}}
        & \makecell{~~~~\small{epoch}}
\end{tabular}
}
% \caption{\small Certified radius $R_t$ along time steps}\label{tab:cert-rad}
\end{subtable}

}

\caption{\small Label prediction accuracy of pseudo label predictor against target models of different overfitting levels on CelebA. We plot label prediction accuracy w.r.t. the number of training epochs of the target model.}
\label{fig:overfit}

\end{figure}

In Fig.~\ref{fig:overfit} we plot label prediction accuracy w.r.t. different training epochs of the target model.
The results indicate that target models of higher overfitted levels are more vulnerable to \method pseudo label predictor.
% \fan{more explanations}
% Intuitively, target models of higher overfitting levels remember more training-set-specific information.
\looseness=-1
In the ideal case, the classification model learns a general mapping from face images to identity labels such that the non-sensitive information will be ignored.
However, in reality, the correlation between the non-sensitive region and ground truth label grows stronger as the number of training epochs increases, as currently the overfitting problem has not been completely solved in machine learning.

\renewcommand{\thesubfigure}{\alph{subfigure}}

\subsection{More Qualitative Results}
\begin{figure}[t]
    \centering
    \includegraphics[width=0.65\textwidth]{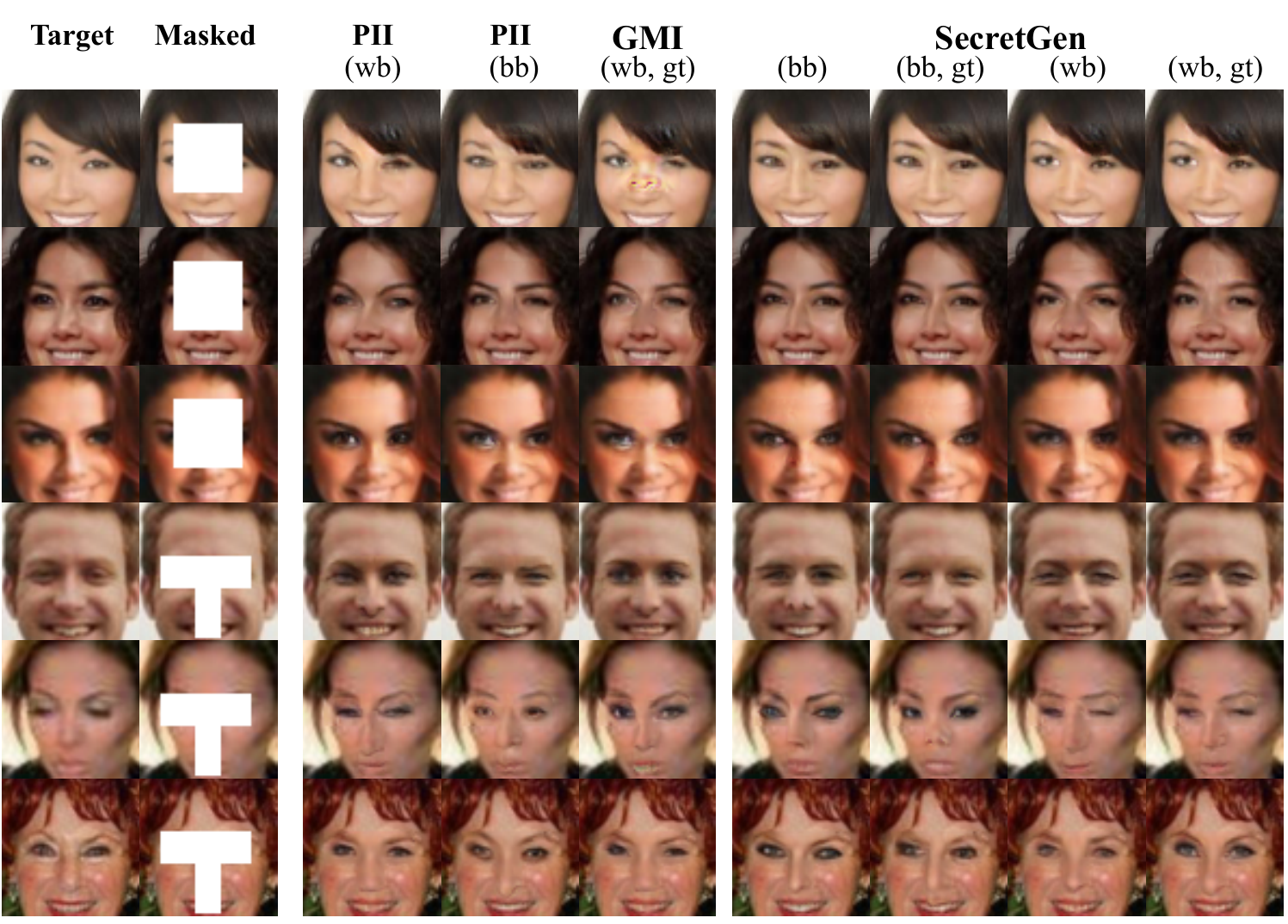}
    \caption{\small More qualitative results on CelebA. ``bb''/``wb'' indicates that the method requires \textit{blackbox}/\textit{whitebox} access to the model. ``gt'' indicates that the method requires ground truth labels. Images in the first column are ground truth private images. Images in the second column are prior information available to the adversary.}
    \label{fig:more_qual}
    
\end{figure}

We present some qualitative results of \method in Fig.~\ref{fig:more_qual}. We can see that when both ground truth label and whitebox access are available, \method can produce recoveries that are closer to the target private image than GMI.
In other settings, \method also outperforms the baseline PII in reconstructing privacy-sensitive attributes of the target image.
Besides, \method generates visually plausible images under all settings.

\label{appx:visual-results}

\end{document}